\documentclass{article}

\usepackage{microtype}
\usepackage{graphicx}
\usepackage{booktabs} 
\usepackage{pdfpages}
\usepackage{caption}
\usepackage{tabularx}
\usepackage{subcaption}
\usepackage{booktabs} 
\usepackage{longtable}
\usepackage{pdflscape}
\usepackage{algorithm}
\usepackage{algorithmicx}
\usepackage{algpseudocode}
\usepackage{hyperref}

\usepackage[accepted]{icml2024}

\usepackage{amsmath}
\usepackage{amssymb}
\usepackage{mathtools}
\usepackage{amsthm}
\usepackage{todonotes}
\usepackage{color} 
\usepackage{amsfonts}
\usepackage{booktabs}
\usepackage{xtab}
\usepackage{afterpage}
\usepackage{booktabs}
\usepackage{multirow}
\usepackage[capitalize,noabbrev]{cleveref}

\theoremstyle{plain}
\newtheorem{theorem}{Theorem}[section]

\theoremstyle{definition}
\newtheorem{definition}[theorem]{Definition}

\theoremstyle{remark}

\icmltitlerunning{Open-Vocabulary Calibration for Fine-tuned CLIP}

\begin{document}

\twocolumn[
\icmltitle{Open-Vocabulary Calibration for Fine-tuned CLIP}

\icmlsetsymbol{equal}{*}
\icmlsetsymbol{workdown}{†}
\begin{icmlauthorlist}
\icmlauthor{Shuoyuan Wang}{sust,um,workdown}
\icmlauthor{Jindong Wang}{wm}
\icmlauthor{Guoqing Wang}{uestc}
\icmlauthor{Bob Zhang}{um}
\icmlauthor{Kaiyang Zhou}{hkbu}
\icmlauthor{Hongxin Wei}{sust}
\end{icmlauthorlist}

\icmlaffiliation{sust}{Department of Statistics and Data Science, Southern University of Science and Technology, Shenzhen, China.}
\icmlaffiliation{um}{Department of Computer and Information Science, University of Macau, Taipa, Macau}
\icmlaffiliation{uestc}{School of Computer Science and Engineering, University of Electronic Science and Technology of China, China. }
\icmlaffiliation{wm}{William \& Mary, Williamsburg, Virginia, USA}
\icmlaffiliation{hkbu}{Department of Computer Science, Hong Kong Baptist University, Hong Kong SAR, China}

\icmlcorrespondingauthor{Hongxin Wei}{weihx@sustech.edu.cn}

\icmlkeywords{Machine Learning, ICML}

\vskip 0.3in
]



\printAffiliationsAndNotice{\icmlWorkDone}  

\begin{abstract}
Vision-language models (VLMs) have emerged as formidable tools, showing their strong capability in handling various open-vocabulary tasks in image recognition, text-driven visual content generation, and visual chatbots, to name a few. In recent years, considerable efforts and resources have been devoted to adaptation methods for improving the downstream performance of VLMs, particularly on parameter-efficient fine-tuning methods like prompt learning. However, a crucial aspect that has been largely overlooked is the confidence calibration problem in fine-tuned VLMs, which could greatly reduce reliability when deploying such models in the real world. This paper bridges the gap by systematically investigating the confidence calibration problem in the context of prompt learning and reveals that existing calibration methods are insufficient to address the problem, especially in the open-vocabulary setting. To solve the problem, we present a simple and effective approach called Distance-Aware Calibration (DAC), which is based on scaling the temperature using as guidance the distance between predicted text labels and base classes. The experiments with 7 distinct prompt learning methods applied across 11 diverse downstream datasets demonstrate the effectiveness of DAC, which achieves high efficacy without sacrificing the inference speed. Our code is available at \href{https://github.com/ml-stat-Sustech/CLIP_Calibration}{https://github.com/ml-stat-Sustech/CLIP\_Calibration}.
\end{abstract}

\section{Introduction}
Vision-language models (VLMs) like CLIP~\cite{radford2021learning} have achieved strong performance in open-vocabulary image recognition by leveraging natural language supervision~\cite{radford2021learning}. To strengthen performance for VLMs in downstream applications, numerous fine-tuning methods have been studied in the literature, among which prompt learning~\cite{zhou2022learning} has gained the most attention due to its parameter-efficient design and robustness. Ideally, in an open-vocabulary setting a fine-tuned VLM should make accurate and reliable predictions on both seen (base) and unseen (novel) classes. While prompt learning algorithms can achieve considerable improvement in accuracy, the reliability issue is still underexplored, preventing the deployment of such models in high-stakes applications like medical diagnosis and autonomous driving.

Existing research has shown that the pre-trained CLIP model is well-calibrated in the zero-shot setting~\cite{minderer2021revisiting}. However, after fine-tuning for downstream tasks, the model struggles with miscalibration, i.e., the predicted class probabilities do not align with the true correctness likelihood~\cite{oh2023towards}. Existing research has only studied the calibration of fine-tuned CLIP on base classes. In this work, we present the first study on calibrating fine-tuned VLMs under an open-vocabulary setting. Our experiments reveal that fine-tuned VLMs exhibit overconfidence in novel classes but underconfidence in base classes. Moreover, we find that the current state-of-the-art post-hoc calibration methods can mitigate the miscalibration problem on base classes. However, the calibration result fails to transfer to novel classes, which motivates us to design post-hoc methods for open-vocabulary calibration.

In this work, we show that the miscalibration issue on novel classes can be mitigated through a simple fix to temperature scaling \cite{guo2017calibration}---the most popular method for post-hoc calibration---by scaling the temperature with class information. Our approach, \textit{Distance-Aware Calibration} (\textbf{DAC}), is motivated by our analysis on the deviation degree of novel classes. Specifically, we show that prompt tuning leads to a large gap between the text features of base classes and those of novel classes. Furthermore, as the deviation degree of a novel class increases, the model becomes more ``confident'' in its predictions.

Concretely, the key idea behind DAC is to scale the temperature value based on the distance between the text embeddings of a novel class and those of base classes. This can be achieved by estimating a textual deviation score for each novel class, which quantifies the deviation degree of the corresponding normalized text features. With the class-dependent scaling factors, our method generates high temperatures for those novel classes far away from the base classes, improving the calibration performance under open-vocabulary settings.

We evaluate our approach DAC using 7 distinct prompt learning methods applied across 11 diverse downstream datasets. The results show that DAC consistently improves open-vocabulary calibration on all prompt learning methods. For example, DAC achieves an average of 6.84\% reduction in Expected Calibration Error (ECE) for CoOp~\cite{zhou2022learning} over the 11 datasets, with maximum reduction reaching 16\%. The same improvement can be achieved for other prompt learning methods as well, such as MaPLe~\cite{khattak2023maple} and PromptSRC~\cite{khattak2023self}. Beyond prompt tuning methods, we also show that DAC can boost the open-vocabulary calibration performance for existing post-hoc methods such as Density-Ratio Calibration~\cite{xiong2023proximity}.

As a post-hoc calibration method, DAC can be easily integrated with various prompt tuning methods in a plug-and-play manner. Our method is computationally efficient and does not require hyperparameter tuning (See Figure~\ref{fig_k}). In short, the main contributions of this paper are summarized as follows:
\begin{itemize}
    \item \textbf{Problem}: We find that fine-tuned VLMs often suffer from miscalibration, and existing post-hoc calibration methods often fail in the open-vocabulary setting.
    \item \textbf{Analysis}: We empirically study the correlation between the calibration and the textual distribution gap. We show that after prompt learning, VLMs tend to be overconfident on classes far away from base classes.
    \item \textbf{Method}: We introduce DAC, a simple and effective post-hoc calibration method that rectifies the predicted confidence while maintaining classification accuracy.
    \item \textbf{Evaluation}: Extensive experiments with 7 tuning VLM tuning methods applied across 11 datasets demonstrate the effectiveness of our approach.
\end{itemize}

\section{Background}

\textbf{Contrastive Language-Image Pretraining (CLIP).}
CLIP is a vision-language model that measures the alignment between image and text, which enables zero-shot inference for open-vocabulary classes \cite{radford2021learning}. Let $\phi$ and $\theta$ denote CLIP's image and text encoders, respectively. Given an image instance $\boldsymbol{x} \in \mathbb{R}^d$ and a text label $c$, the logit function of CLIP can be formulated as:
\begin{equation}
\label{eq1}
L_{c}^{clip} \left ( \boldsymbol{x}_{i} \right ) =\tau \cdot \operatorname{sim}\left(\phi(\boldsymbol{x}), \psi(\boldsymbol{t}_c)\right),
\end{equation}
where $\phi$ and $\psi$ denote the image and text encoders in CLIP. $\boldsymbol{t}_c$ is derived from a hand-crafted prompt like ``a photo of a \{class\}", where the ``\{class\}" is filled with the text label $c$. $\tau$ is generally set as a constant of 100.

For multi-class classification, we predict by selecting the label with the highest probabilities among the label candidate set $\mathcal{C} = \{c_i\}^K_{i=1}$, as shown below:
\begin{equation}
\label{eq:pred_prob}
c^*=\underset{c \in \mathcal{C}}{\arg \max}\ p(c|\boldsymbol{x})
= \underset{c \in \mathcal{C}}{\arg \max}\ \frac{e^{L_{c}^{clip}\left(\boldsymbol{x}\right)}}{\sum_{i=1}^{K} e^{L_{i}^{clip}\left(\boldsymbol{x}\right)}}
\end{equation}
where $p(c|\boldsymbol{x})$ is the predicted probability of class $c$ for the instance $\boldsymbol{x}$.

\paragraph{Prompt tuning.} To strengthen the CLIP model in downstream applications, prompt tuning algorithms are normally designed to optimize the context prompt \cite{zhou2022learning, zhou2022conditional} for high efficiency. 

In particular, CoOp \cite{zhou2022learning} replace the hand-crafted textual tokens with a set of learnable textual token $\mathcal{T}=\left\{\boldsymbol{v}_{1}, \boldsymbol{v}_{2}, \ldots, \boldsymbol{v}_{M}\right\}$, where $M$ is the length of tokens. 
Then, the logit function is:
\begin{equation}
\label{eq_fs}
L_{c}^{tuned} \left ( \boldsymbol{x} \right ) =\tau \cdot \operatorname{sim}\left(\phi(\boldsymbol{x}), \psi(\boldsymbol{t}^{\prime}_c)\right),
\end{equation}
where $\boldsymbol{t}_{c}^{\prime} = \left [ \boldsymbol{v}_{1}, \boldsymbol{v}_{2},\boldsymbol{v}_{3},...\boldsymbol{v}_{M},\boldsymbol{c} \right ]$ and $\boldsymbol{c}$ denotes the textual embedding of class $c$. During tuning, the learnable textual tokens $\mathcal{T}$ are optimized to minimize the cross-entropy loss $\ell_{ce}$ with labeled few-shot samples $\mathcal{D}=\{(\boldsymbol{x}_i, c_i)\}^N_{i=1}$. Formally, the optimization of CoOp is:
$$
\mathcal{T}^{\star} = \underset{\mathcal{T}}{\arg\min} \frac{1}{N}\sum_{i=1}^{N} \ell_{ce}(p(c_i|\boldsymbol{x}_i))
$$
To maintain the generalization on the unseen classes within the same task, i.e., open-vocabulary classes, CoCoOp \cite{zhou2022conditional} employs instance-conditional prompts $\boldsymbol{t}_m(\boldsymbol{x}) = \boldsymbol{t}_m + h_{\theta}(\boldsymbol{x})$,
where $h_{\theta}(\cdot)$ is a neural network parameterized by $\theta$ and $m\in \{1,2,\ldots, M\}$.
While the tuning methods achieve promising performance, the reliability of the prediction probability $p(c|\boldsymbol{x})$ is still a mystery for tuned VLMs. We proceed by introducing the calibration in the prompt tuning setting.

\paragraph{Confidence calibration.} For multi-class classification tasks, vision-language models are designed to produce the class probabilities $p(c|\boldsymbol{x})$ as defined in Equation~\ref{eq:pred_prob}. In addition to high accuracy, it is generally expected for vision-language models to be well calibrated, i.e., the predicted class probabilities can faithfully estimate the true probabilities of correctness. Formally, a \textit{perfectly} calibrated model satisfies $Pr(c^\star = c\mid p(c|\boldsymbol{x})=q) = q, \forall q \in [0,1]$.

To quantify the degree of miscalibration, the \textit{Expected Calibration Error} (ECE) is defined as the difference between accuracy and confidence. With $N$ samples grouped into $K$ bins $\{b_1, b_2, \ldots, b_{K}\}$, the ECE is calculated as:
\begin{equation}
    \mathrm{ECE}=\sum_{k=1}^{K} \frac{\left|b_{k}\right|}{N}\left|\operatorname{acc}\left(b_k\right)-\operatorname{conf}\left(b_k\right)\right|,
\end{equation}
where $\operatorname{acc}\left(\cdot\right)$ and $\operatorname{conf}\left(\cdot\right)$ denotes the average accuracy and confidence in bin $b_{k}$.

\section{An Empirical Study}


In the literature, it has been shown that pre-trained CLIP is well calibrated in zero-shot inference \cite{minderer2021revisiting}. 
In this section, we empirically study the calibration performance of tuned VLMs.  In particular, we compare the zero-shot CLIP \cite{radford2021learning} with a variety of popular prompt tuning algorithms, including: 
CoOp \cite{zhou2022learning}, CoCoOp \cite{zhou2022conditional}, MaPLe \cite{khattak2023maple}, and ProGrad \cite{zhu2023prompt}.

\begin{figure}[t]
  \centering
  \includegraphics[width=0.48\textwidth]{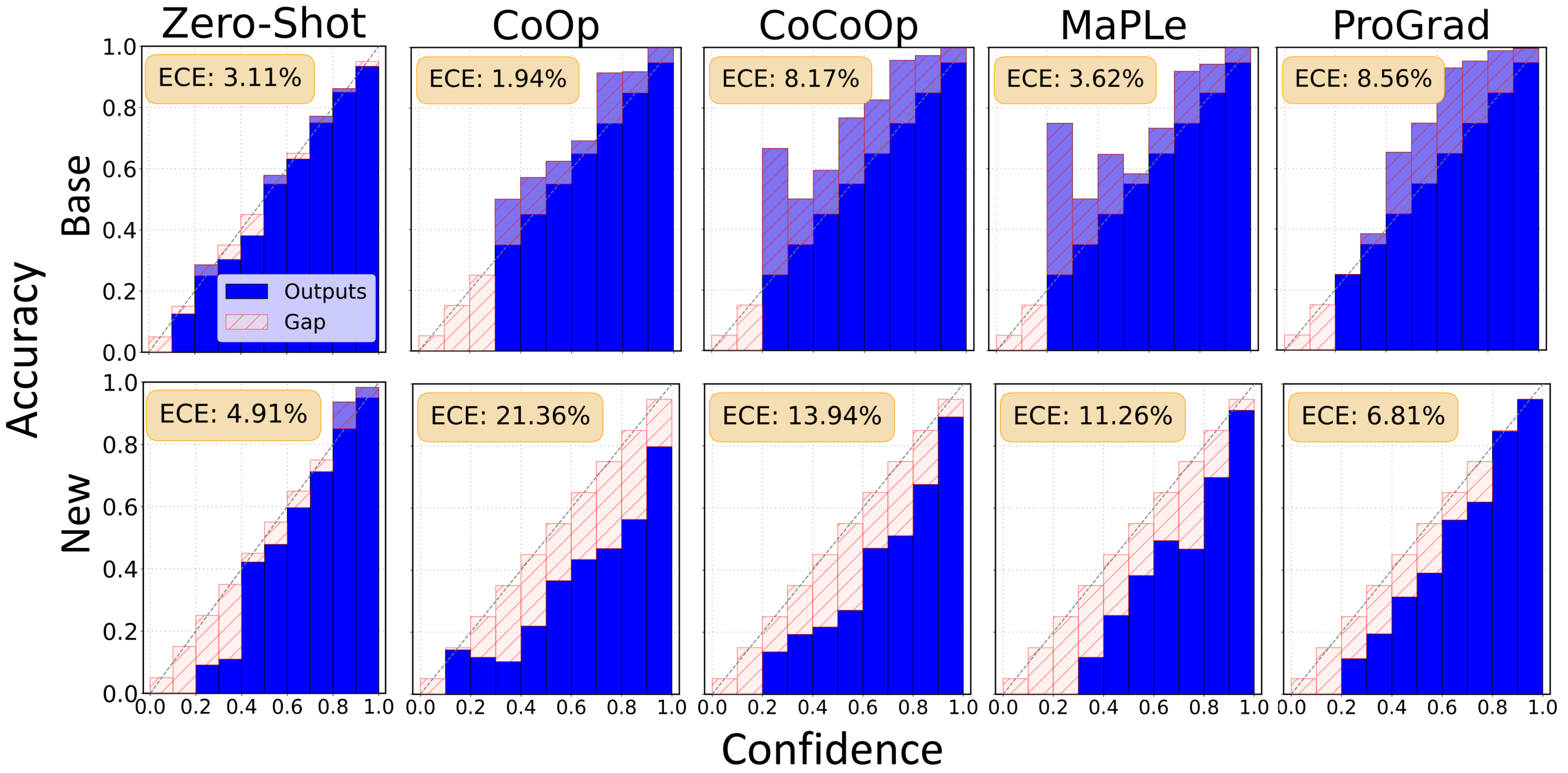}
  \caption{Reliability of fine-tuned CLIP (ViT-B/16) on the Flower102 dataset. ECE: Expected Calibration Error (lower is better). Miscalibration is depicted in pink for overconfidence and purple for underconfidence.}
\label{fig_fs_ece}
\end{figure}

\subsection{Are fine-tuned VLMs well-calibrated?}
To showcase the miscalibration in tuned VLMs, we fine-tune the pre-trained CLIP
with 7 tuning methods on 11 downstream datasets and use ECE as the calibration metric. 
The descriptions of compared methods and datasets can be found in Subsection \ref{sec_exp_setup}. 
The miscalibration of tuned CLIP on the Flower102 dataset is visualized in Figure \ref{fig_fs_ece} through reliability diagrams. 
Intuitively, one might expect that deep models will demonstrate increased confidence in the learned base classes after tuning and tend to make more conservative predictions under open-vocabulary conditions since they are not explicitly learned during the tuning process.
On the contrary, we make a counter-intuitive observation that \textit{tuned VLMs tend to be underconfident in base classes while prone to overconfident in new classes.}
We present the detailed calibration results in Appendix \ref{calibration_fs}.




\subsection{Can fine-tuned VLMs be calibrated?}
\label{sec_miscab_vlm}

To address the above miscalibration in tuned CLIP, we utilize commonly-used post-hoc confidence calibration for tuned VLMs and expect to calibrate the model well on both base and new classes. 
Specifically, post-hoc calibration can be split into two categories: \textit{scaling-based methods} and \textit{binning-based methods}. Specifically, regarding the former, we use Temperature Scaling (TS) \cite{guo2017calibration} and Density-Ratio Calibration (DEN) \cite{xiong2023proximity}. 
For the latter, we use Histogram Binning (\cite{zadrozny2001obtaining}), Isotonic Regression (IR) \cite{zadrozny2002transforming} and Multi-Isotonic Regression (MIR) \cite{zhang2020mix}.
The complete calibration results are available in Appendix \ref{appx_fs_cab}.
The calibration results on ImageNet-1K are shown in Table \ref{tab_fs_cab}, and we can make the following findings:

(1) \textbf{Post-hoc calibration can remedy miscalibration in base classes.} 
As a complement of the observation in the recent works \cite{tu2023closer, oh2023towards}, we observe that both scaling-based methods ( e.g., DEN) and bin-based approaches ( e.g., MIR) can effectively reduce miscalibration in base classes. 
(2) \textbf{Post-hoc calibration on base classes can not transfer to new classes.}  
For instance, the calibration performance of ``TS" and ``DEN" is worse than the model without calibration (``Conf") or tuning (``ZS"). This phenomenon indicates that simply using the calibrator learned in base classes is not optimal for wider unseen distributions \cite{yu2022robust}.
For bin-based methods, they need the probabilities from base classes as the input, which are incompatible with zero-shot prediction.
This limitation leads to an open question: \textit{How to calibrate tuned VLMs in open-vocabulary classes?} 

\begin{table}[t]
  \centering
  \caption{ECE (\%) of fine-tuned CLIP with different calibration methods. We use ProDA to fine-tune CLIP-ViT-B/16 on ImageNet-1K. ``ZS'' means zero-shot CLIP and ``Conf'' means confidence score without calibration after tuning. ``–'' means the results are not applicable. ``Conf'' shows underconfidence in base classes. ``TS'' and ``DEN'' show overconfidence in new classes.}
    \resizebox{0.48\textwidth}{!}{
    \begin{tabular}{cccccccc}
    \toprule
     & ZS    & Conf  & TS    & DEN & HB    & IR    & MIR \\
    \midrule
    Base classes  & 3.58  & 4.82  & 1.94  & 0.73  & 4.23  & 2.09  & 0.82  \\
    New classes   & 2.09  & 1.59  & 3.90  & 3.86  & -     & -     & - \\
    \bottomrule
    
    \end{tabular}%
    \label{tab_fs_cab}%
    }
\end{table}%



\section{Open-Vocabulary Calibration}
In this section, we first investigate the open-vocabulary calibration with a visualization of the joint vision-language embedding space and show that the textual gap between the base and new classes is crucial for calibration.
Next, we quantify this gap by a distance-based metric for the deviation degree of a novel class. 
We then present our method and its advantages for open-vocabulary calibration.

\subsection{Feature Space Analysis}
In the prompt tuning of VLMs, the primary influence of tuning is reflected in the textual features.
This fact motivates us to explore how these changes in text features correlate to the miscalibration in the new classes.
To understand the open-vocabulary miscalibration in VLMs after prompt tuning, we first analyze it from the joint vision-language representation space. 
Specifically, following \textit{modality gap}\footnote{https://github.com/Weixin-Liang/Modality-Gap/} \cite{liang2022mind}, paired image-text of the downstream datasets are fed into zero-shot / tuned CLIP and their embeddings are visualized in 2D using singular value decomposition (SVD). 

As mentioned in  \cite{liang2022mind}, we observe that images and text are embedded at arm’s length in the shared representation space. 
In zero-shot CLIP, the embeddings of the same modality are relatively concentrated. 
However, as is shown in Figure \ref{fig_coop_svd}, there is a distinct textual distribution gap between the base and new classes after tuning. 
Furthermore, ECE has been significantly increased, indicating that the model had been miscalibrated. 
Informally, we hypothesize that \textit{the deviation degree in the textual gap is crucial for open-vocabulary calibration.}


\subsection{Motivation}
\label{sec_motivation}
To verify the hypothesis, we first introduce a distance-based metric \textit{proximity} to quantify the deviation degree of features.

\begin{definition}[Proximity \cite{xiong2023proximity}] Consider a feature $\mathbf{z}\in \mathbb{R}^{d}$ as the embedding of a test sample and the held-out feature embeddings $\mathcal{E} \in \mathbb{R}^{n \times d}$, proximity is a function inversely correlates with the mean distance between the test sample and its $K$ nearest neighbors in held-out sets:
\begin{equation}
P(\mathbf{z}, \mathcal{E}) =  \sigma \left(\frac{1}{K} \sum_{\mathbf{x}_i \in \mathcal{N}_K(\mathbf{z}, \mathcal{E})} \operatorname{dist}(\mathbf{z}, \mathbf{x}_i)\right),
\end{equation}
\end{definition}
where $\mathcal{N}_K(\mathbf{z}, \mathcal{E})$ denotes the set of $K$ nearest neighbors of $\mathbf{z}$ and $\operatorname{dist}\left(\cdot, \cdot\right)$ is a predefined metric for distance calculation
$\sigma$ is a non-negative, monotonically decreasing function.
This definition implies a greater likelihood that the test sample originates from the same distribution as the held-out set as its proximity increases. 

In VLMs, we have a set of ${l}_{2}$-normalized \cite{radford2021learning} textual embeddings of seen class  $\mathcal{W} = \left \{\boldsymbol{w}_{j}\right \} _{j=1}^{S}$. Given an arbitrary class $c_{i}\in \mathcal{U}$ in the test time and its corresponding class token is $\boldsymbol{c}_{i}$. We first obtain its normalized textual embedding $\boldsymbol{w}_{i}=\theta\left(\boldsymbol{t}_{i}\right)$, where $\boldsymbol{t}_{i} = \left \{  \boldsymbol{v}_{1}, \boldsymbol{v}_{2},\boldsymbol{v}_{3},...\boldsymbol{v}_{M},\boldsymbol{c}_{i} \right \}$ represents the textual token. we can define the textual distribution gap between this class  $c_{i}$ and the seen class as follows:
\begin{equation}
    \label{eq:proximity}
    P(\boldsymbol{w}_{i}, \mathcal{W})=\exp \left(-\frac{1}{K} \sum_{\boldsymbol{w}_{j} \in \mathcal{N}_{K}(\boldsymbol{w}_{i}, \mathcal{W} )} \|\boldsymbol{w}_{i} - \boldsymbol{w}_{j}\|_2\right).
\end{equation}
Here  we use $e^{-x}$ as $\sigma\left ( \cdot  \right ) $ and  ${l}_{2}$-distance for $\operatorname{dist}\left(\cdot, \cdot\right)$. 

\begin{figure}[t]
    \centering
    \begin{subfigure}{.235\textwidth}
        \centering
        \includegraphics[width=\linewidth]{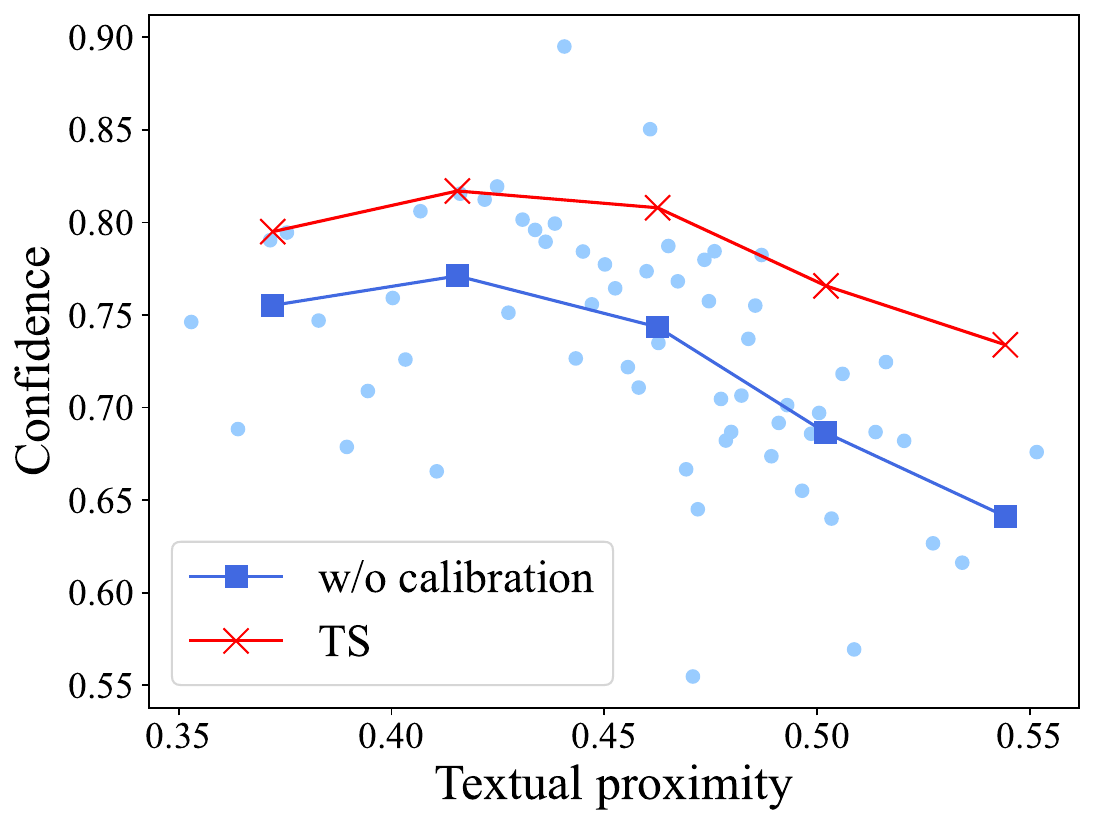}
        \caption{Confidence}
        \label{fig_cls_wise_conf}
    \end{subfigure}
    \hfill
    \begin{subfigure}{.235\textwidth}
        \centering
        \includegraphics[width=\linewidth]{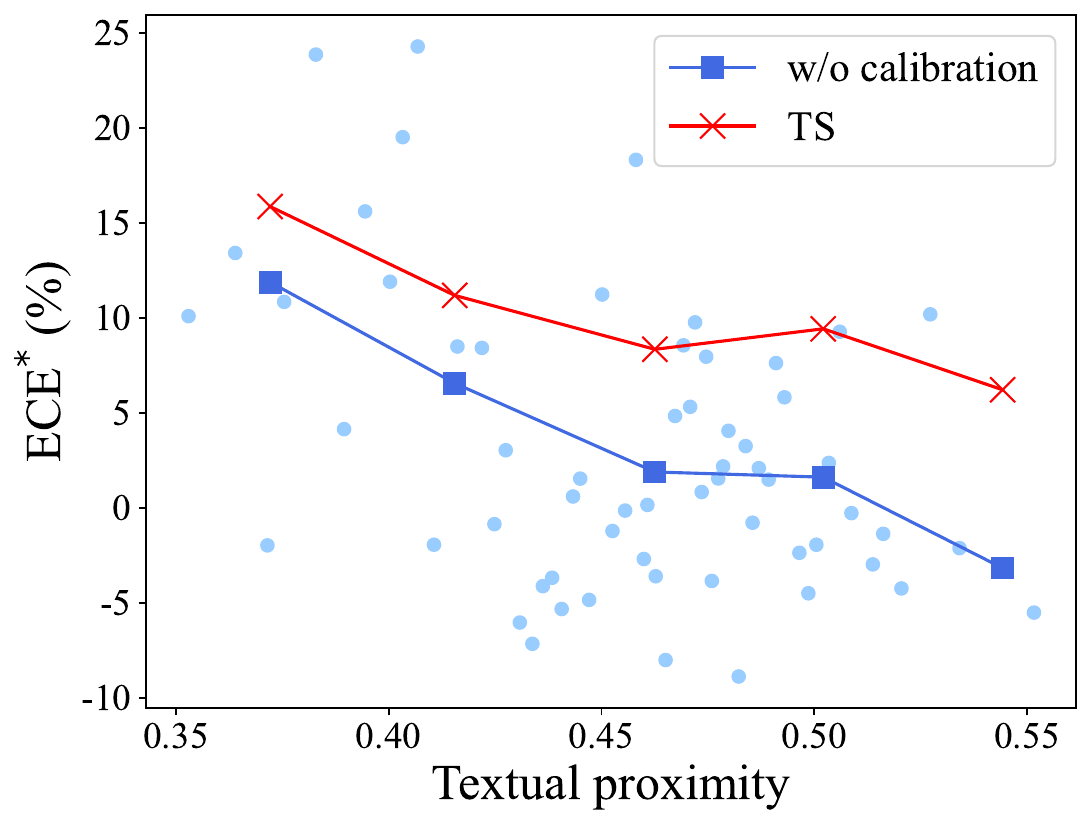}
        \caption{$\text{ECE}^{*}$}
        \label{fig_cls_wise_ece}
    \label{fig_cls_ece_std}
    \end{subfigure}
    \caption{Class-wise performance on StanfordCars dataset after tuning. $\text{ECE}^{*}$ with a positive (negative) value denotes overconfidence (underconfidence). The scatters represent the origin results and the broken line denotes the bin-based results. Confidence and $\text{ECE}^{*}$ increase as proximity decreases. Temperature scaling (TS) can not mitigate the overconfidence.}
    \label{fig_cls_wise_proximity}
\end{figure}

We present the correlation between the deviation degree of textual feature and calibration in Figure \ref{fig_cls_wise_proximity}. For each textual embedding of new classes, we can calculate its corresponding proximity to the hold-out set of base textual features. 
We can observe that:

\begin{figure}[t]
    \centering
    \begin{subfigure}{.23\textwidth}
        \centering
        \includegraphics[width=\linewidth]{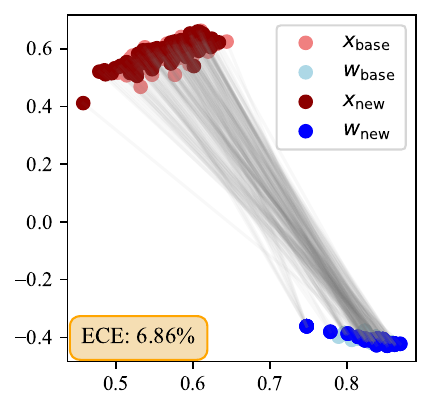}
        \caption{Zero-Shot}
        \label{fig_zs_svd}
    \end{subfigure}
    \hfill
    \begin{subfigure}{.23\textwidth}
        \centering
        \includegraphics[width=\linewidth]{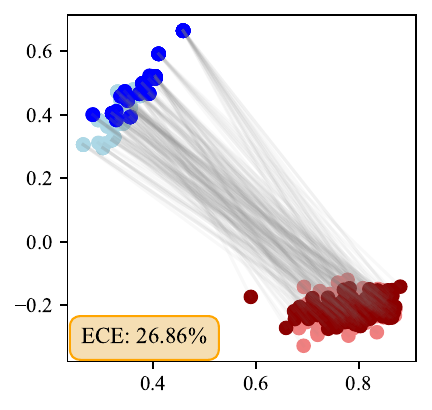}
        \caption{CoOp}
        \label{fig_coop_svd}
    \end{subfigure}
    \caption{Paired inputs from image ($x$) / text ($w$) are sampled from the DTD dataset fed into zero-shot / tuned CLIP and are visualized in 2D using SVD. Compared with zero-shot CLIP, CoOp has a larger textual distribution gap between the base and new classes}
    \label{fig_svd}
\end{figure}

\paragraph{Lower proximity correlates with higher confidence and ECE.} 
Specifically, if the given text is far from the base textual distribution, tuned VLM tends to present a higher confidence, which results in worse calibration performance. 
Similar to the pre-training stage \cite{liang2022mind}, contrastive loss in V-L tuning bridges a connection between visual and textual modality. 
Each modality will be embedded into a relatively concentrated distribution.
However, in the test time, if the given class is an outlier from the textual distribution learned in tuning, tuned CLIP  may not be closely aligned across modalities and fail to give a proper predicted confidence level. 
Furthermore, as shown in Figure \ref{fig_cls_ece_std}, commonly used calibration techniques such as Temperature Scaling (TS) do not alleviate but deteriorate this issue. 
We present more details of the experiment are included in Appendix \ref{sec_textual_proximity}.
This motivated us to modify the confidence level when the prediction exhibits lower textual proximity. 



\subsection{Distance-Aware Calibration}

Per the analysis above, the predicted class with lower textual proximity can result in higher confidence and worse calibration performance.
In this work, we propose a post-hoc method to rectify the predicted confidence level.
More specifically, we introduce a textual deviation-based score to scale the temperature value, thereby updating the confidence level to align with the true correctness likelihood.

\paragraph{Textual deviation estimation.} 
Ideally, the model is expected to give highly uncertain predictions for examples from novel classes, with relatively low accuracy. With this in mind, we define the score by calculating the deviation distances of novel classes to base classes using tuned CLIP models, compared to those of pre-trained CLIP.

Let $\boldsymbol{w}_i$ and $\boldsymbol{w}_{i}^{\prime}$ be the normalized text features of class $c_i$ from the pre-trained and tuned VLMs respectively. The \textbf{T}extual \textbf{D}eviation (\textbf{TD}) score for class $c_i$ is formulated as: 
\begin{equation}
 \gamma(c_i) = \frac{ P(\boldsymbol{w}_{i}^{\prime}, \mathcal{W}^{\prime})}{P(\boldsymbol{w}_{i}, \mathcal{W})},
\end{equation}


where   $ \mathcal{W} $ and $ \mathcal{W}^{\prime} $ denote the sets of normalized textual features for base classes. 
For example, a low TD score indicates that the corresponding class is more likely to be deviant for the textual distribution of base classes. As demonstrated in Subsection~\ref{sec_motivation}, tuned VLMs tend to be overconfident in examples from such classes. Thus, we propose to update the confidence level according to the TD score for each class.

\paragraph{Calibrated inference.} Before evaluation, we first collect the TD scores  $\Gamma = \left  \{ q_{i} \right \}_{i=1}^{K}$ for each class. Given a test instance $\boldsymbol{x}$, we first calculate its predicted class $\hat{c}=\arg \max _{c} p(c \mid \boldsymbol{x})$ using the tuned CLIP model. Finally, the rectified logit is calculated as :
\begin{equation}
L_{c}^{dac} \left ( \boldsymbol{x} \right ) =\gamma(\hat{c}) \cdot \tau \cdot \operatorname{sim}\left(\phi(\boldsymbol{x}), \psi(\boldsymbol{t}^{\prime}_c)\right).
\end{equation}
where $\boldsymbol{t}^{\prime}_{c}$ is the textual token of class $c$, defined in Equation~\ref{eq_fs}.
Here, we rectify the confidence level of the prediction by scaling the temperature value $\tau$ with TD score $\gamma(\hat{c})$.  
For base classes, we set $\gamma_{\hat{c}} = 1$, and our DAC is degraded to vanilla temperature scaling. Thus, our method would not affect the calibration performance on base classes.
Noticeably, DAC offers several compelling advantages:

\begin{itemize}

    \item \textbf{Realistic}: The calculation procedure of DAC only utilizes the information of text labels. Thus, our method does not assume access to the visual information from open-vocabulary concepts, which is challenging to obtain in real applications.

    \item \textbf{Algorithm-agnostic}: DAC can easily combine with existing prompt tuning algorithms, consistently improving the calibration performance in novel classes. 


    \item \textbf{Easy to use}: our method does not require hyperparameter tuning (see Table~\ref{fig_k}) and does not introduce extra computational cost (See Table~\ref{tab_image_text_proximity}).
\end{itemize}



\section{Experiments}
\label{sec:experiment}
In this section, we verify the effectiveness of our proposed DAC with 11 downstream classification benchmark datasets.

\begin{table*}[t]
\caption{Average calibration performance across 11 datasets. ``Conf'' represents the origin performance on open-vocabulary classes with existing tuning methods. ``DAC'' to our method applied to existing tuning methods. ↓ indicates smaller values are better. Calibration error is given by $\times 10^{-2}$. \textbf{Bold} numbers are significantly superior results.}
\centering
\renewcommand\arraystretch{1.2}
\begin{tabular}{lccccccccccc}
\toprule
             & \multicolumn{2}{c}{ECE(↓)} &  & \multicolumn{2}{c}{ACE(↓)} &  & \multicolumn{2}{c}{MCE(↓)} &  & \multicolumn{2}{c}{PIECE(↓)} \\ \cline{2-3} \cline{5-6} \cline{8-9} \cline{11-12} 
Method       & Conf        & DAC           &  & Conf         & DAC           &  & Conf        & DAC            &  & Conf          & DAC            \\ \midrule
CoOp         & 13.84    & \textbf{7.00}   &  & 13.76    & \textbf{6.91}   &  & 3.80    & \textbf{1.71}    &  & 14.71     & \textbf{9.02}    \\
CoCoOp       & 6.29     & \textbf{4.82}   &  & 6.21     & \textbf{4.77}   &  & 1.79    & \textbf{1.40}    &  & 8.07      & \textbf{7.15}    \\
ProDA        & 4.27     & \textbf{3.99}   &  & 4.35     & \textbf{4.08}   &  & 1.27    & 1.32             &  & 6.57      & \textbf{6.35}    \\
KgCoOp       & 4.36     & 4.32   &  & 4.43     & 4.38   &  & 1.18    & 1.13    &  & 6.67      & 6.63    \\
MaPLe        & 5.77     & \textbf{4.61}   &  & 5.71     & \textbf{4.64}   &  & 1.82    & \textbf{1.42}    &  & 7.59      & \textbf{6.98}    \\
ProGrad      & 4.22     & \textbf{3.74}   &  & 4.27     & \textbf{3.74}   &  & 1.22    & \textbf{1.09}    &  & 6.75      & \textbf{6.55}    \\
PromptSRC    & 3.84     & \textbf{3.63}   &  & 3.92     & \textbf{3.69}   &  & 1.09    & 1.08    &  & 6.26      & 6.17    \\ 
\bottomrule
\end{tabular}
\label{tab_main_exp}
\end{table*}

\subsection{Experimental Setup}
\label{sec_exp_setup}

\paragraph{Benchmark setting.} We evaluate the confidence calibration of existing popular prompt tuning methods. Following the few-shot evaluation protocol \cite{zhou2022conditional}, the datasets are split into base and new classes. The model is trained only on the base classes in a few-shot setting and we generally report the calibration performance on new classes.


\paragraph{Datasets.} In this work, we follow the standard base-to-new setting \cite{zhou2022learning, zhou2022conditional} and use 11 image recognition datasets in our experiment.
The datasets cover diverse classification tasks. It includes general object datasets like ImageNet \cite{deng2009imagenet} and Caltech101 \cite{fei2004learning}, alongside fine-grained classification datasets such as OxfordPets \cite{parkhi2012cats}, StanfordCars \cite{krause20133d}, Flowers102 \cite{nilsback2008automated}, Food101 \cite{bossard2014food}, and FGVCAircraft \cite{maji2013fine} . For specialized categories, it features SUN397 \cite{xiao2010sun} for scene recognition, UCF101 \cite{soomro2012ucf101} for action recognition, DTD \cite{cimpoi2014describing} for texture classification, and EuroSAT \cite{helber2019eurosat} with its satellite images.

\paragraph{Compared methods.} We compare and incorporate our method with existing VLM tuning algorithms, which can be split into two categories: \textit{prompt tuning} and \textit{multi-modal tuning}: For prompt tuning, we use Context Optimization (CoOp) \cite{zhou2022learning}, Conditional Context Optimization (CoCoOp) \cite{zhou2022conditional}, Prompt Distribution Learning (ProDA) \cite{lu2022prompt}, Knowledge-guided Context Optimization (KgCoOp) \cite{yao2023visual}, Prompt-aligned Gradient (ProGrad) \cite{zhu2023prompt}. 
For multi-modal tuning, we use Multi-modal Prompt Learning (MaPLe) \cite{khattak2023maple},  Prompting with Self-regulating Constraints (PromptSRC) \cite{khattak2023self}.

\paragraph{Evaluation metrics.} 4 standard metrics are used in our evaluation of open-vocabulary confidence calibration: Expected Calibration Error (ECE) \cite{guo2017calibration},  Maximum Calibration Error (MCE) \cite{guo2017calibration}, Adaptive Calibration Error (ACE) \cite{nixon2019measuring} and Proximity-Informed Expected Calibration Error \cite{xiong2023proximity}.

\begin{figure}[t]
    \centering
    \begin{subfigure}{.235\textwidth}
        \centering
        \includegraphics[width=\linewidth]{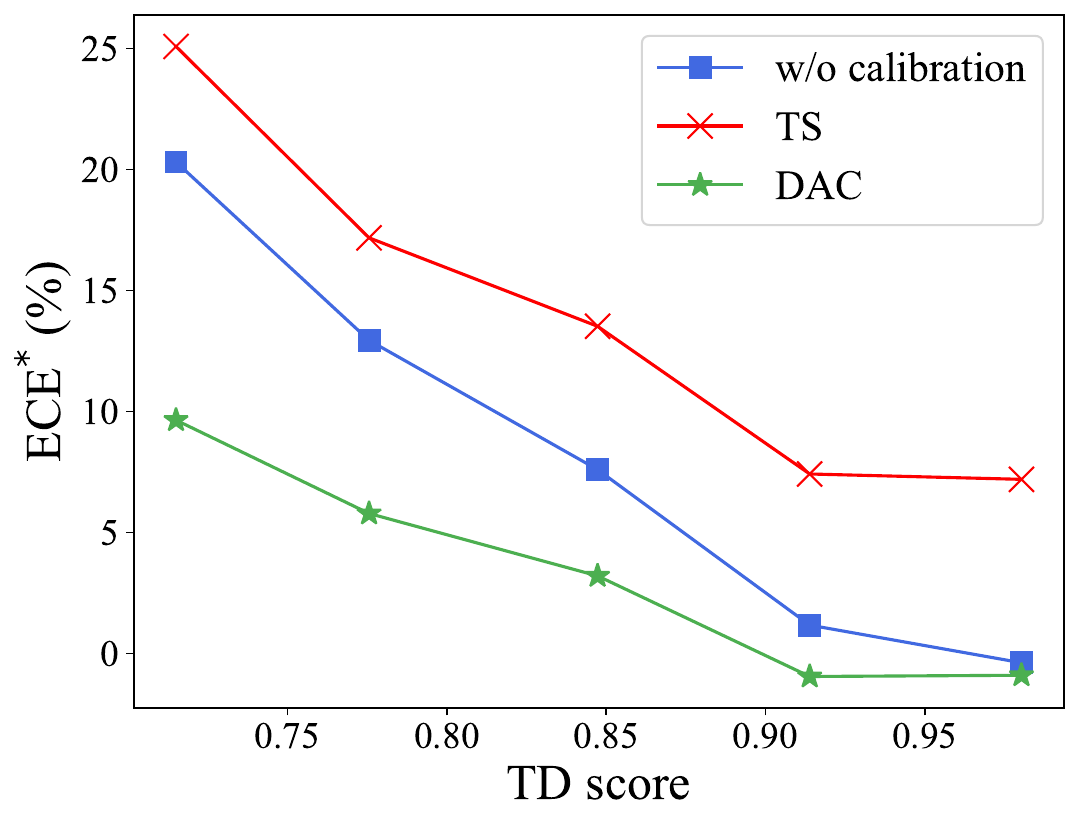}
        \caption{StanfordCars}
        \label{subfig_td_cars}
    \end{subfigure}
    \hfill
    \begin{subfigure}{.235\textwidth}
        \centering
        \includegraphics[width=\linewidth]{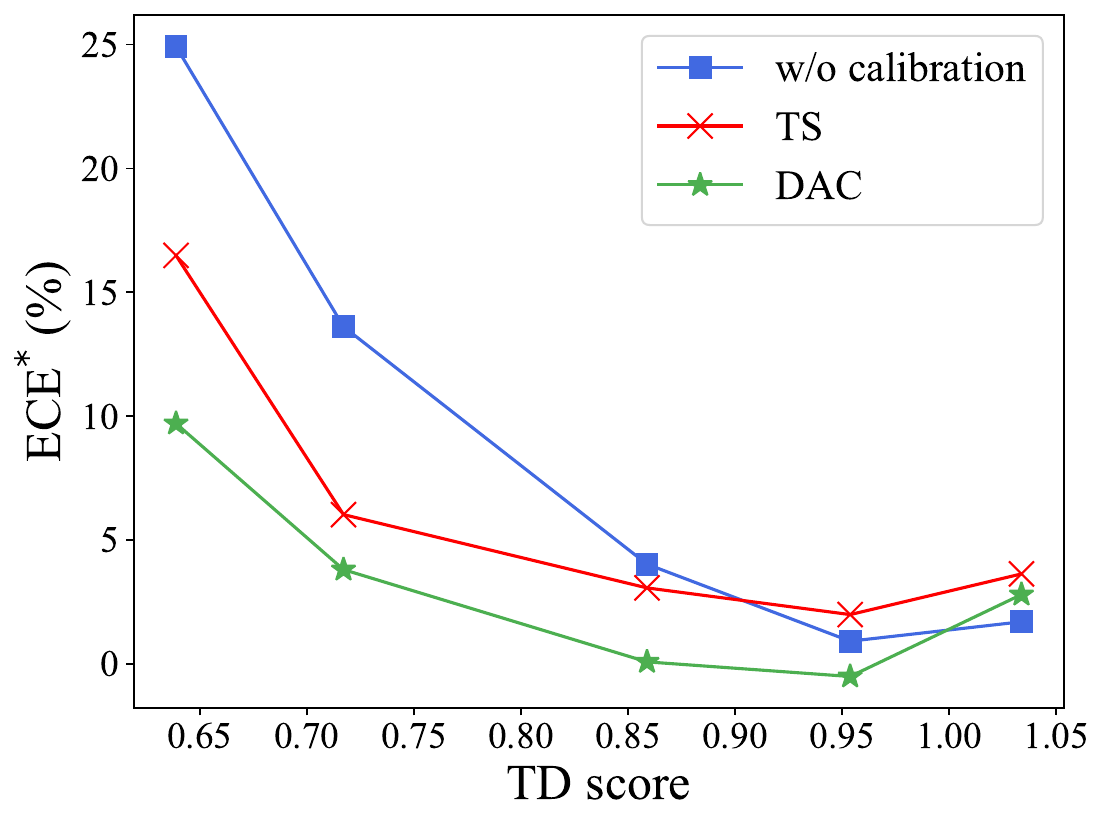}
        \caption{SUN397}
        \label{subfig_td_sun}
    \end{subfigure}
    \caption{$\text{ECE}^{*}$ (\%) performance with difference calibration methods. $\text{ECE}^{*}$ with a positive (negative) value denotes overconfidence (underconfidence). Our proposed DAC largely mitigates the overconfidence in the predicted classes with lower TD score.}
    \label{fig_td_score}
\end{figure}

\textbf{Implementation details.} For the main results, We use CLIP ( ViT-B/16) \cite{radford2021learning} as the pre-trained VLM throughout our experiments and report results averaged over 3 runs. We fine-tune the model with 16 samples per class in a few-shot setting \cite{zhou2022conditional}. For the compared tuning methods, we adopt them from the corresponding official implementation. For our proposed DAC, for simplicity, we set the number of nearest textual features $K$=5 throughout the paper to calculate TD scores. More details on datasets, pre-trained models, hyperparameters, and implementation details can be found in Appendix \ref{sec_exp_setting}.

\begin{figure}[t]
    \centering
    \begin{subfigure}{.235\textwidth}
        \centering
        \includegraphics[width=\linewidth]{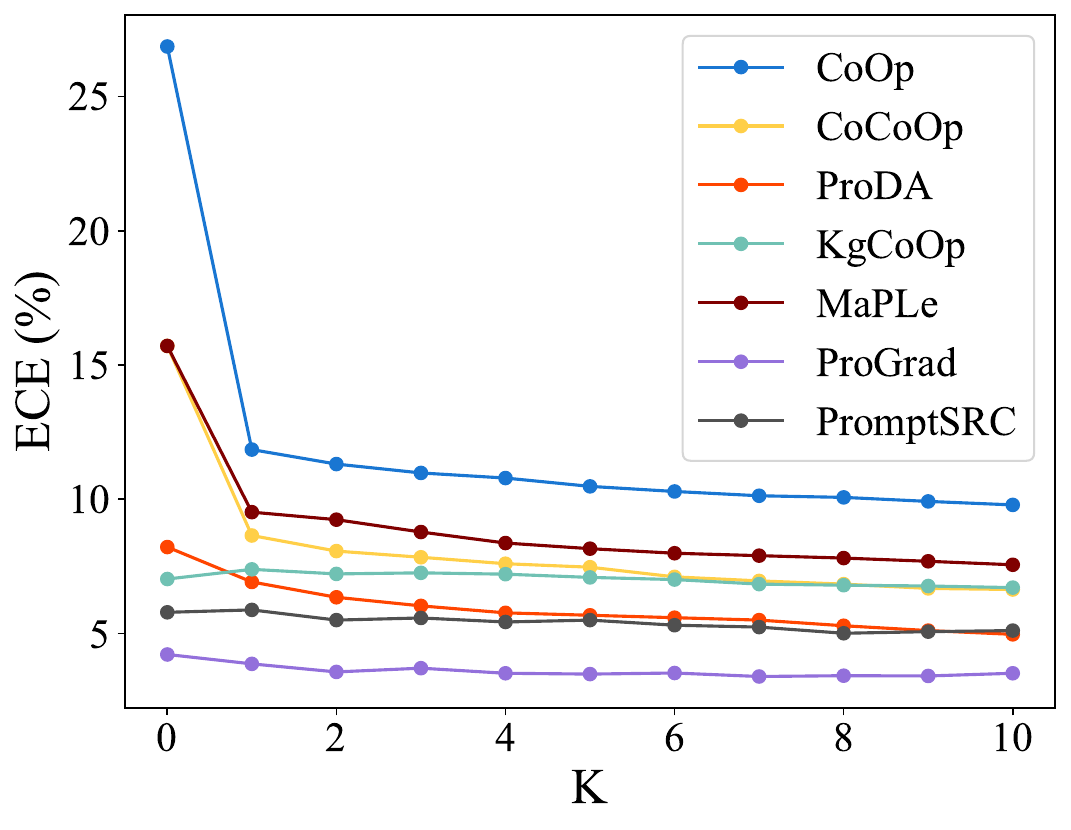}
        \caption{DTD}
        \label{subfig_k_dtd}
    \end{subfigure}
    \hfill
    \begin{subfigure}{.235\textwidth}
        \centering
        \includegraphics[width=\linewidth]{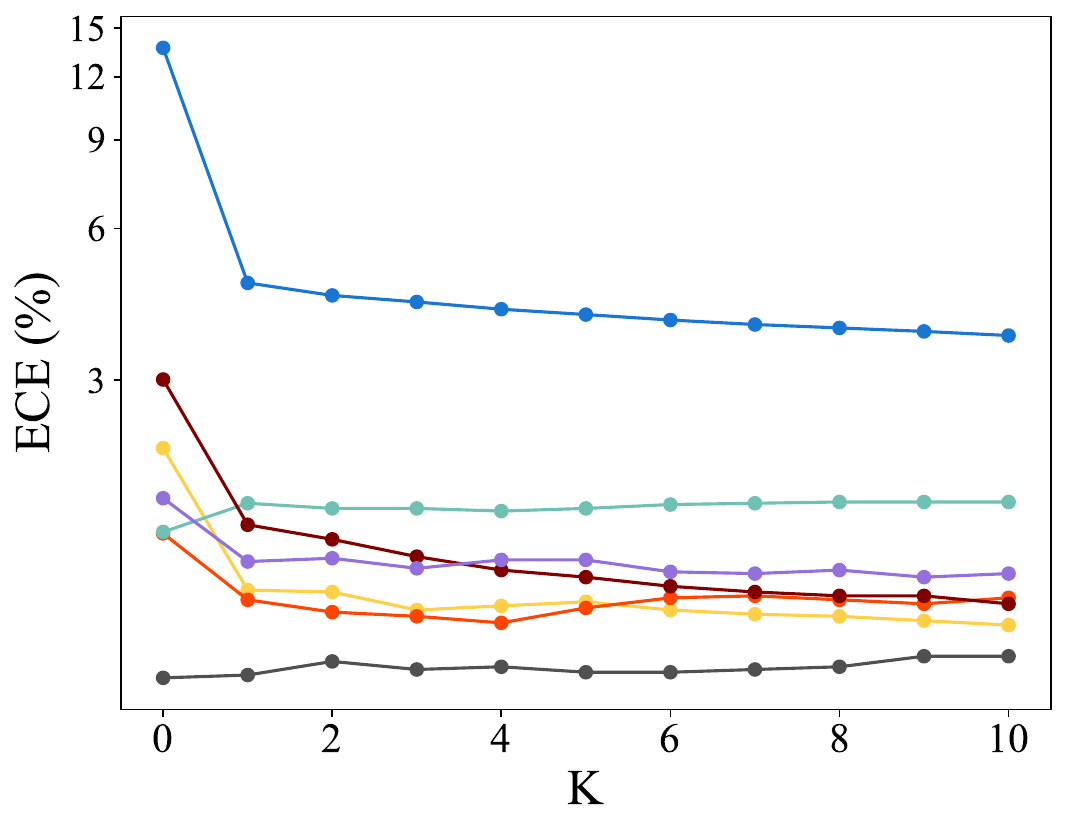}
        \caption{SUN397}
        \label{subfig_k_sun}
    \end{subfigure}
    \caption{Hyperparameter sensitivity of the number of neighbors used in computing textual proximity. The miscalibration will be noticeably mitigated if $K>1$.}
    \label{fig_k}
\end{figure}

\subsection{Results}

\begin{table*}[h]
\caption{Calibration results of ECE (\%) across different confidence levels. $\Delta$ shows the improvement achieved by DAC. \textbf{Bold} numbers denote the top-3 most significant improvements.}
\centering
 \resizebox{0.9\textwidth}{!}{
\begin{tabular}{lccccccccccc}
\toprule
Method &  & 0.1 & 0.2 & 0.3 & 0.4 & 0.5 & 0.6 & 0.7 & 0.8 & 0.9 & 1.0 \\
\midrule
     & Conf & 0.00 & 19.24 & 18.86 & 15.87 & 20.42 & 33.28 & 30.30 & 37.84 & 41.60 & 18.57 \\
CoOp & DAC & 0.00 & 4.95 & 8.17 & 11.33 & 6.51 & 11.42 & 24.12 & 17.41 & 11.37 & -0.94 \\
     & $\Delta$ & 0.00 & -14.29 & -10.69 & -4.54 & -13.91 & \textbf{-21.86} & -6.18 & \textbf{-20.43} & \textbf{-30.23} & -19.51 \\
\midrule
      & Conf & 0.00 & 0.00 & -12.80 & 3.90 & 16.73 & 10.50 & 38.07 & 23.93 & 19.11 & 12.13 \\
MaPLe & +DAC & 0.00 & -3.62 & -15.32 & 5.72 & 6.74 & 3.12 & 15.45 & 6.16 & 9.29 & 6.55 \\
      & $\Delta$ & 0.00 & -3.62 & -2.52 & 1.82 & \textbf{-10.99} & -7.38 & \textbf{-22.62} & \textbf{-17.77} & -9.82 & -5.58 \\
\midrule
& Conf & 0.00 & -3.82 & 0.14 & -0.10 & 4.29 & 6.31 & 3.48 & 8.11 & 1.23 & 4.86 \\
ProGrad & +DAC & 0.00 & -0.71 & 0.03 & 1.30 & -1.32 & 0.40 & -0.65 & -0.04 & 0.44 & -0.34 \\
& $\Delta$ & 0.00 & 3.11 & -0.11 & 1.40 & \textbf{-5.61} & \textbf{-5.91} & -4.13 & \textbf{-8.15} & -0.79 & -5.20 \\
\bottomrule
\end{tabular}
}
\label{tab_bin}
\end{table*}

\begin{table}[t]
\centering
\caption{Comparison results of ECE (\%) using different visual backbones on Flower102 dataset. The smaller values are better.}
 \resizebox{0.50\textwidth}{!}{
\begin{tabular}{ccccccccc}
\toprule
& \multicolumn{2}{c}{CoOp} &       & \multicolumn{2}{c}{CoCoOp} &       & \multicolumn{2}{c}{ProGrad} \\
\cline{2-3} \cline{5-6} \cline{8-9} 
Backbone & Conf   & DAC           &  & Conf     & DAC           &  & Conf     & DAC            \\ \hline
RN50     & 15.72  & \textbf{8.03}   &  & 6.00      & \textbf{4.88}   &  & 4.1      & \textbf{3.39}    \\
ViT-B-32 & 21.07  & \textbf{11.72}  &  & 9.71     & \textbf{6.57}   &  & 5.11     & \textbf{4.36}    \\
ViT-B-16 & 18.34  & \textbf{10.19}  &  & 11.49    & \textbf{7.74}   &  & 5.45     & \textbf{5.04}   \\
\bottomrule
\end{tabular}
}
\label{tab_backbone}
\end{table}

\textbf{DAC improves open-vocabulary calibration in existing prompt tuning.} 
Table \ref{tab_main_exp} shows the average calibration error of all datasets.
For the open-vocabulary calibration performance, our proposed DAC effectively mitigates the miscalibration in various tuning methods across all metrics. 
For instance, DAC significantly reduces the ECE from 13.84\% to 7.00\%, which makes CoOp more reliable in terms of predicted confidence.
Furthermore, note that DAC is also applicable to \textit{multi-modal tuning} (e.g., MaPLe and PromptSRC). 
This approach adjusts both vision and language branches and DAC consistently improves the open-vocabulary calibration performance of these methods.
The results indicate that the textual distribution gap widely exists in various tuning methods and our proposed DAC can consistently boost open-vocabulary calibration upon existing tuning methods. 
Due to space constraints, we provide detailed calibration results in Appendix \ref{appx_main_exp}.

\textbf{DAC significantly reduces calibration error, especially for high-confidence predictions.} To verify where DAC helps the most over the baseline methods, we analyzed the calibration performance across various confidence levels. Specifically, we measure the gap between confidence and accuracy for each bin on the DTD dataset in Table \ref{tab_bin}. A negative gap indicates underconfident prediction, and vice versa. The difference between the baseline and our method is presented for easy comparison of improvements in each bin. The results show that DAC improves the calibration performance mainly in those confident predictions (the bins 0.6-0.9). In particular, our method can significantly alleviate the overconfidence issue. For example, in the case of CoOp, our method reduces the gap from 41.60 to 11.37, a direct improvement of 30.23, in the bin with confidence 0.9.

\textbf{DAC regularizes the prediction with low TD score for better calibration.}
Since DAC adjusts the temperature according to the predicted class, we present a detailed visualization of class-wise results in \ref{fig_td_score}.
For a comprehensive evaluation, we also compare our results with Temperature Scaling \cite{guo2017calibration}.
As the TD score decreases, the tuned model shows higher confidence in open-vocabulary classes, and TS cannot help and even exacerbates the issue. 
In particular, DAC addresses this overconfidence by reducing the logit scale for the prediction with low textual proximity, thereby ensuring more reliable predictions.
We present the detailed results in the Appendix \ref{sec_td_score}.

\textbf{DAC is robust to the number of nearest neighbors.}
We evaluate how $K$ in our method affects the open-vocabulary calibration performance. 
Specifically, we present this ablation with ECE based on DTD and SUN397 datasets.
We vary the number of textual neighbors $k=\{1,2, 3,...,10\}$.
As is shown in Figure \ref{fig_k}, an increase in the number of neighbors beyond 0 leads to an evident reduction in ECE, and the performance starts to reach a point of saturation with the further addition of neighbors.
It's noteworthy that DAC shows robustness to the choice of $K$, even if we set $K=1$ yielding significant calibration improvements.
This phenomenon indicates that it is beneficial to adjust the confidence level according to the textual distribution gap between base and new classes.
For simplicity, we employ a moderate range of neighbors and use $K=5$ throughout our experiments.


\begin{figure}[t]
    \centering
    \begin{subfigure}{.235\textwidth}
        \centering
        \includegraphics[width=\linewidth]{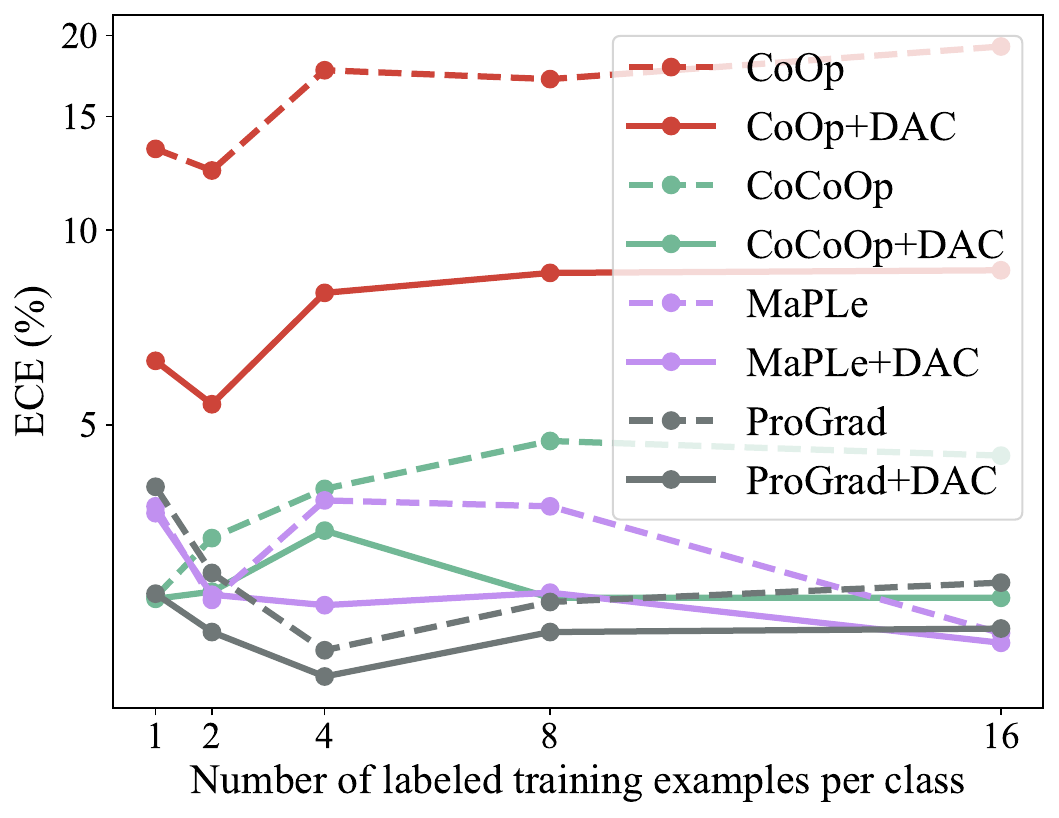}
        \caption{UCF101}
        \label{subfig_shot_dtd}
    \end{subfigure}
    \hfill
    \begin{subfigure}{.235\textwidth}
        \centering
        \includegraphics[width=\linewidth]{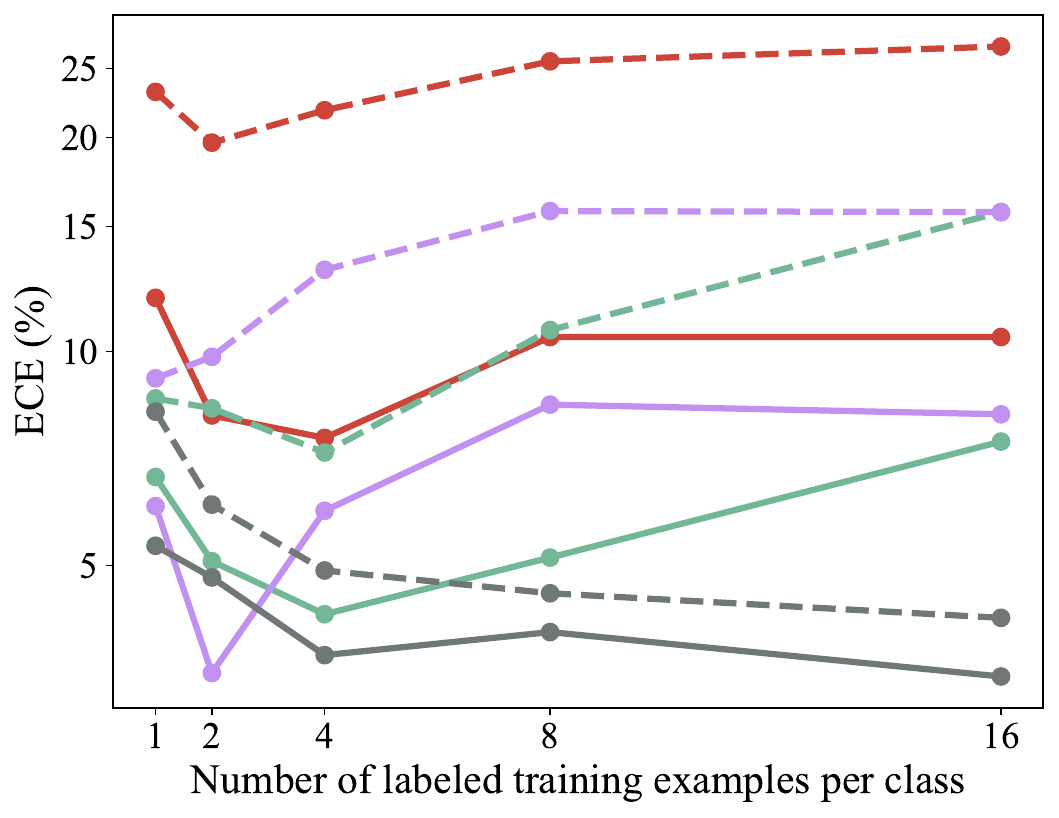}
        \caption{DTD}
        \label{subfig_shot_sun}
    \end{subfigure}
    \caption{Comparison results of ECE (\%) using different shots. Miscalibration is a common issue and DAC can reduce it across different shots. The Y-axis is presented in an exponential form for a better view.
}
    \label{fig_shot}
\end{figure}


\textbf{DAC is effective across various few-shot settings.} 
We consider two options in few-shot learning: shots and visual backbone, where "shot" is the number of labeled samples per class in tuning.
Unlike accuracy, which tends to increase as the number of shots increases \cite{zhou2022learning}, calibration errors are prevalent across diverse few-shot settings.
We visualize the ECE performance across different shots in Figure \ref{fig_shot} and demonstrate that our proposed DAC successfully mitigates this issue.
Moreover, we show that DAC is effective across a variety of backbones in Table \ref{tab_backbone}.


\section{Discussion}

\textbf{How does DAC perform in full fine-tuning?}
To evaluate the flexibility of DAC, we provide additional results with a most recent fine-tuning approach named FLYP\cite{goyal2023finetune}, where both image and text encoders will be updated. Following the official implementation, we fine-tune the model for 10 epochs with a learning rate of 1e-5.

As is shown in Table \ref{tab_flyp},  DAC successfully reduces the calibration error of models finetuned by FLYP on open-vocabulary classes, which is consistent with the results of prompt-tuning methods. Furthermore, we observe that fine-tuning methods like FLYP also tend to overfit the base classes on several datasets, making it more challenging to calibrate on open-vocabulary classes.

\begin{table}[t]
\caption{Calibration of CoOp and FLYP using ECE (\%) across 11 datasets. \textbf{Bold} numbers are significantly superior results.}
\centering
 \resizebox{0.45\textwidth}{!}{
\begin{tabular}{lcccc}
\toprule
 & \multicolumn{2}{c}{CoOp} & \multicolumn{2}{c}{FLYP} \\
\cmidrule(lr){2-3} \cmidrule(lr){4-5}
Dataset & Conf & +DAC & Conf & +DAC \\
\midrule
Caltech101 & 4.08 & \textbf{3.17} & 2.80 & \textbf{2.01} \\
OxfordPets & 1.83 & 1.82 & 0.73 & 0.51 \\
StanfordCars & 12.50 & \textbf{5.16} & 19.90 & \textbf{15.62} \\
Flowers102 & 18.34 & \textbf{10.19} & 20.58 & \textbf{19.09} \\
Food101 & 3.83 & \textbf{1.78} & 5.74 & \textbf{4.13} \\
FGVCAircraft & 28.44 & \textbf{17.38} & 36.52 & \textbf{26.84} \\
SUN397 & 13.70 & \textbf{4.05} & 21.72 & \textbf{19.09} \\
DTD & 26.86 & \textbf{10.48} & 12.30 & \textbf{6.49} \\
EuroSAT & 12.71 & \textbf{8.62} & 5.32 & 5.30 \\
UCF101 & 19.24 & \textbf{8.67} & 11.98 & \textbf{8.18} \\
ImageNet & 10.69 & \textbf{5.67} & 8.42 & \textbf{7.19} \\
\midrule
AVG & 13.84 & \textbf{7.00} & 13.27 & \textbf{10.40} \\
\bottomrule
\end{tabular}
}
\label{tab_flyp}
\end{table}

\textbf{Visual proximity vs. Textual proximity.}
Proximity plays an important role in trustworthy machine learning \cite{jiang2018trust, liu2023simple, xiong2023proximity, yuksekgonul2023beyond}.  
Recent works mainly focus on the proximity estimation of the model input and largely mitigate the miscalibration in supervised learning.
However, we show that textual proximity is more important than visual proximity in prompt tuning.
To this end, we compare the calibration results and inference efficiency of our proposed DAC on ImageNet-1K dataset with Density-Ratio Calibration \cite{xiong2023proximity}.

We present the results in Table \ref{tab_image_text_proximity}.
For calibration performance, our proposed DAC outperforms DEN in the calibration of new classes, and we can achieve better calibration results by combining DAC with DEN.
Additionally, we assess the efficiency of inference. 
In terms of computational cost, our approach maintains high efficiency with negligible extra computational burden compared to DEN and the baseline (Conf). 
Overall,  our proposed DAC shows performance superiority and inference efficiency on calibration in new classes compared with methods based on image proximity.


\begin{table}[t]
  \centering
  \caption{Calibration performance and inference time on ImageNet-1K. CoOp is used for prompt tuning. ``base'' denotes the performance in base classes. DAC shows performance superiority and inference efficiency on calibration in new classes.}
    \begin{tabular}{ccccc}
    \toprule
          & Conf  & DAC  & DEN  & DAC+DEN \\
    \midrule
    ECE\textsubscript{base} & 1.65  & 1.65  & 1.19  & 1.19  \\
    ECE\textsubscript{new} & 10.69 & 5.67  & 8.08  & 3.29  \\ \midrule
    time (s) & 0.12  & 0.13  & 15.88 & 15.91 \\
    \bottomrule
    \end{tabular}%
  \label{tab_image_text_proximity}%
\end{table}%

\textbf{Textual feature normalization is critical.}
In DAC, we normalize the features from the textual encoder for calculating TD scores. 
However, the significance of such normalization is not immediately evident. In this ablation, we contrast the performance of textual proximity-based calibration with and without feature normalization. 
As an illustration, we report the calibration results of tuned CLIP with CoOp.

\begin{figure}[t]
  \centering
  \includegraphics[width=0.48\textwidth]{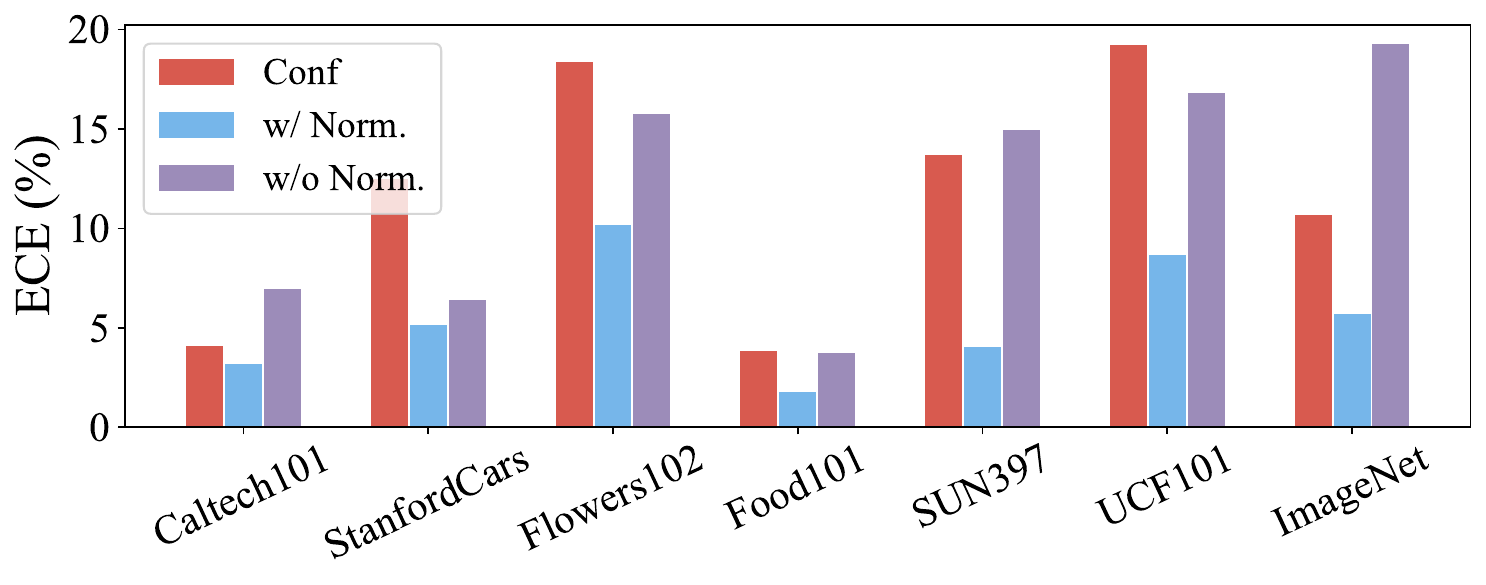}
  \caption{Ablation results of textual feature normalization with CoOp. We compare the effect of using normalization in the textual feature vs. without normalization.} 
\label{fig_text_normalization}
\end{figure}

Our results in Figure \ref{fig_text_normalization} show that employing normalized features can have a better estimation of class-wise proximity, which achieves better calibration performance than raw features. 
Previous works \cite{jiang2024diverse, sun2022out, xiong2023proximity} demonstrated that feature norms may misestimate the K-nearest-neighbor-based score with Euclidean distance.
Empirically, we verify that normalization plays a key role in our distance-based calibration.


\textbf{Calibration with data from novel classes.}
Generally, it is impractical for a user to access the image-text pairs from the novel classes during tuning. As an analysis, we verify the feasibility of open-vocabulary calibration utilizing the knowledge from new classes for post-hoc calibration. 
To this end, We present the calibration results that the model will be tuned in base classes and calibrated using open-vocabulary image-text pairs, which utilize Temperature Scaling (TS) as the post-hoc calibration method.

As illustrated in Table \ref{tab_ts_new},  TS effectively calibrates the model using image-text pairs of the new classes.
However, such calibration compromises the model's ability to deliver a reliable confidence level for the base classes.
Given that TS modifies the logits scale with a single scalar, it is infeasible to calibrate the model on both base and new class.
Overall, as a training-free calibration, our proposed DAC boosts the calibration performance according to textual proximity.

\section{Related work}
\label{sec_relatedwork}


\textbf{Prompt tuning.}
Thanks to the rich natural language supervision, pre-trained large vision-language models \cite{jia2021scaling, radford2021learning} can effectively comprehend visual concepts and apply them in downstream tasks (e.g. image classification \cite{radford2021learning, zhou2022learning, lu2022prompt, naeem2023i2mvformer}, image captioning \cite{hu2022scaling, wang2023image, wang2023controllable}, visual question answering \cite{parelli2023clip}) in a zero-shot manner.
Despite VLM's effectiveness in generalizing new visual concepts, the performance of zero-shot VLM still lags behind the fine-tuned performance on specific downstream tasks \cite{zhang2023vision}.
To further boost the downstream adaptation of pre-trained VLMs, many parameter-efficient tuning methods like prompt tuning \cite{zhou2022learning, zhou2022conditional, yao2023visual, zhu2023prompt, zang2022unified} have been proposed for high efficiency. 
Recently, several works have explored multimodel tuning \cite{khattak2023maple, khattak2023self, shen2024multitask}, which adapts both vision and language branches of CLIP for better modality alignment.
Despite the great success of V-L tuning, their effectiveness on safety-related evaluation is less explored.
In this work, we first investigate the calibration performance for fine-tuned VLMs. 
We present that fine-tuned CLIP generally suffers from miscalibration, and our proposed DAC can effectively mitigate the issue by rectifying the logit scale based on textual information.

\textbf{Confidence calibration.}
Confidence calibration has been widely studied to ensure that the confidence levels output by models accurately reflect their true performance. 
To achieve this, A common practice is to employ post-hoc calibration techniques after model training.
Generally, post-hoc calibration can be split into two paradigms: \textit{scaling-based} methods \cite{guo2017calibration, tomani2022parameterized, yu2022robust, zhang2020mix, joy2023sample, xiong2023proximity} and \textit{bin-based} methods \cite{zadrozny2001obtaining, zadrozny2002transforming, zhang2020mix, roelofs2022mitigating}.
For scaling-based calibration, a representative method is temperature scaling \cite{guo2017calibration}, which learns a single scalar for rescaling the logit. 
Mix-n-Match \cite{zhang2020mix} leverages ensemble and composition techniques to achieve data efficiency and maintain accuracy in confidence estimates. 
ATS \cite{joy2023sample} modifies the predicted confidence by per-data-point adaptive temperature.
For binning-based calibration, the samples are segmented into various bins according to confidence levels, with each bin undergoing calibration individually. Recently, several works have explored calibration in VLMs \cite{oh2023towards, levine2023enabling}. 
In this work, we focus on the open-vocabulary calibration of fine-tuned VLMs.
We first show that established post-hoc calibration on base classes can not transfer to new classes.
To this end, our DAC fixes the logit scale of prediction via textual deviation-informed score in new classes.

\begin{table}[t]
  \centering
  \caption{Average performance of ECE (\%) on 11 datasets. CoOp is used to tune the pre-trained CLIP. TS$_{\text{new}}$ / TS$_{\text{base}}$ denotes the calibration on data of base/new classes. }
    \begin{tabular}{ccccc}
    \toprule
    Class & Conf & TS$_{\text{base}}$ & TS$_{\text{new}}$ & DAC \\
    \midrule
    Base  & 2.95  & 2.57  & 13.05  & 2.95  \\
    New   & 13.84  & 12.95  & 7.56  & 7.00  \\
    \bottomrule
    \end{tabular}%
  \label{tab_ts_new}%
\end{table}%

\section{Conclusion}

In this paper, we introduce Distance-Aware Calibration (DAC), a simple fix to temperature scaling in fine-tuned VLMs, that enhances existing prompt tuning methods in open-vocabulary calibration.
By scaling the temperature value according to the distance between the predicted text label and base classes, fine-tuned CLIP tends to give a more reliable confidence level for new classes.
Extensive experiments show that DAC can improve the calibration in new classes after prompt tuning while maintaining the performance in base classes.
Our proposed DAC can be easily incorporated with existing scaling-based calibration without compromising the inference speed.
We hope our findings inspire further research on the calibration of VLMs.

\section*{Acknowledgements}
This research is supported by the Shenzhen Fundamental Research Program (Grant No. JCYJ20230807091809020). Bob Zhang is supported by the Science and Technology Development Fund (FDCT), Macao, SAR, under Grant 0028/2023/RIA1. Guoqing Wang is supported in part by the National Natural Science Foundation of China under grant U23B2011, 62102069, U20B2063 and 62220106008, the Key R\&D Program of Zhejiang under grant 2024SSYS0091. We gratefully acknowledge the support of the Center for Computational Science and Engineering at the Southern University of Science and Technology for our research.

\section*{Impact Statement}
This paper presents work whose goal is to advance the field of Machine Learning. There are many potential societal consequences of our work, none of which we feel must be specifically highlighted here.

\bibliography{reference}
\bibliographystyle{icml2024}

\newpage
\appendix
\onecolumn

\section{Experimental Setting}
\label{sec_exp_setting}


\subsection{Dataset}
Following the base-to-new generalization setting in CoCoOp \cite{zhou2022conditional}, we split the dataset into two subsets \textit{base} and \textit{new}. The labels for these subsets do not overlap.
As is shown in Table \ref{tab_dataset}, we split the dataset into three folds through our experiments.
The pre-trained CLIP will be first tuned on base classes. 
For former calibration methods like \cite{guo2017calibration}, they need a calibration set to align the predicted probabilities more closely with the actual likelihoods of the outcomes. 
Since we can not access new (open-vocabulary) classes,  we sample a calibration set from the datasets of base classes which does not overlap with the training set. 
Finally, we evaluate the model on both base classes and new classes. In our experiments, we mainly focus on the calibration in new classes.

\begin{table}[ht]
  \centering
  \caption{The split strategy of downstream datasets. We use data from base classes to tune and calibrate the pre-trained CLIP.}
    \begin{tabular}{cccc}
    \toprule
    Dataset & $\text{Train}^\dagger$ & $\text{Calibration}^\dagger$ & $\text{Test}^\psi$ \\
    \midrule
    Caltech101 & 800   & 200   & 916 \\
    OxfordPets & 304   & 76    & 1,788 \\
    StanfordCars & 1,568  & 392   & 4039 \\
    Flowers102 & 816   & 204   & 1,410 \\
    Food101 & 816   & 204   & 15,000 \\
    FGVCAircraft & 800   & 200   & 1,667 \\
    SUN397 & 3,184  & 796   & 9,900 \\
    DTD   & 384   & 96    & 828 \\
    EuroSAT & 80    & 20    & 3,900 \\
    UCF101 & 816   & 204   & 1,849 \\
    ImageNet & 8,000  & 25,000 & 25,000 \\
    \bottomrule
    \multicolumn{4}{l}{$\dagger$: Base class, $\psi$: New class.}\\
    \end{tabular}%
  \label{tab_dataset}%
\end{table}%

\subsection{Compared methods}
\label{subsec_tuning_para}
We use CLIP ( ViT-B/16) \cite{radford2021learning} as the pre-trained VLM throughout our experiments and report results averaged over 3 runs. We fine-tune the model with 16 samples per class in a few-shot setting \cite{zhou2022conditional}.
Following the official implementation, We list the general hyperparameters in Table \ref{tab_tuning_hyperparameters}. 
Here, we briefly introduce the corresponding exclusive hyperparameters of each VLM tuning method. All the methods are adopted from their official implementation. For CoOp and CoCoOp, they do not contain other hyperparameters. For ProDA, we set $\lambda = 0.1$. For KgCoOp, we set $\lambda = 8.0$. For MaPLe, we set prompt depth $J$ to 0 and the language and vision prompt lengths to 2. For ProGrad, we set $\lambda = 0.8$. For PromptSRC, we set deep prompting with $V=T=4$. $\lambda_{1}=10$ and $\lambda_{2}=25$ are used in weight loss. For textual diversity, we use a total of $N = 60$ standard prompt templates. 
We reproduce their methods in our framework based on the standard codebase \footnote{https://github.com/KaiyangZhou/CoOp\label{fn_coop}}.

\begin{table}[htbp]
  \centering
  \caption{Hyperparameters for VLM tuning methods. ``BS'' denotes the batch size. ``LR'' denotes the learning rate. ``CTX'' is the context length of the learnable prompt.}
    \begin{tabular}{cccccc}
    \toprule
    Methods & Epochs & BS & LR & CTX \\
    \midrule
    CoOp  & 200   & 32    & 0.002 & 16 \\
    CoCoOp & 10    & 1     & 0.002 & 4 \\
    ProDA & 100   & 4     & 0.001 & 16 \\
    KgCoOp & 200   & 32    & 0.002 & 16 \\
    MaPLe & 5     & 4     & 0.0026 & 2 \\
    ProGrad & 100   & 32    & 0.002 & 16 \\
    PromptSRC & 50    & 4     & 0.0025 & 4 \\
    \bottomrule
    \end{tabular}%
  \label{tab_tuning_hyperparameters}%
\end{table}%

\section{Detailed class-wise results of open-vocabulary calibration}
We present the miscalibration in tuned CLIP and the effectiveness of our proposed DAC via class-wise visualization.

\paragraph{Experiment setup.} 
For tuning methods, we use Context Optimization (CoOp) \cite{zhou2022learning}, Conditional Context Optimization (CoCoOp) \cite{zhou2022conditional}, Knowledge-guided Context Optimization (KgCoOp) \cite{yao2023visual}, Prompt-aligned Gradient (ProGrad) \cite{zhu2023prompt}, Multi-modal Prompt Learning (MaPLe) \cite{khattak2023maple}. 
We calculate their class-wise results according to the \textit{prediction} (pseudo label).
For the datasets, we present the results of various image recognition datasets including ImageNet \cite{deng2009imagenet}, StanfordCars \cite{krause20133d}, Flowers102 \cite{nilsback2008automated}, Food101 \cite{bossard2014food}, and FGVCAircraft \cite{maji2013fine}, SUN397 \cite{xiao2010sun}, UCF101 \cite{soomro2012ucf101} and DTD \cite{cimpoi2014describing}.
For a better view, we visualize bin-base results by broken lines, and the origin results are given by scatters.

\subsection{Textual Proximity}
\label{sec_textual_proximity}
\begin{figure}[h]
    \centering
    \begin{subfigure}{.235\textwidth}
        \centering
        \includegraphics[width=\linewidth]{figs/p/p_StanfordCars_conf.pdf}
        \caption{StanfordCars}
    \end{subfigure}
    \hfill
    \begin{subfigure}{.235\textwidth}
        \centering
        \includegraphics[width=\linewidth]{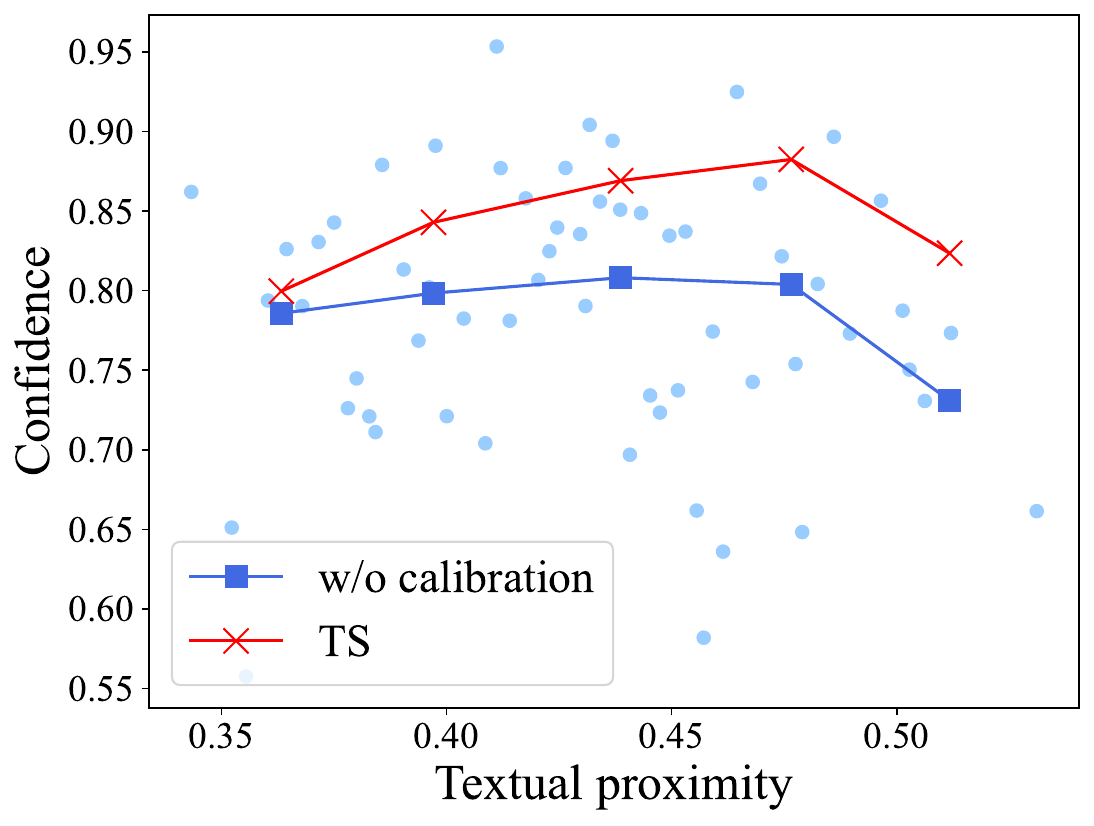}
        \caption{Flowers102}
    \end{subfigure}
    \hfill
    \begin{subfigure}{.235\textwidth}
        \centering
        \includegraphics[width=\linewidth]{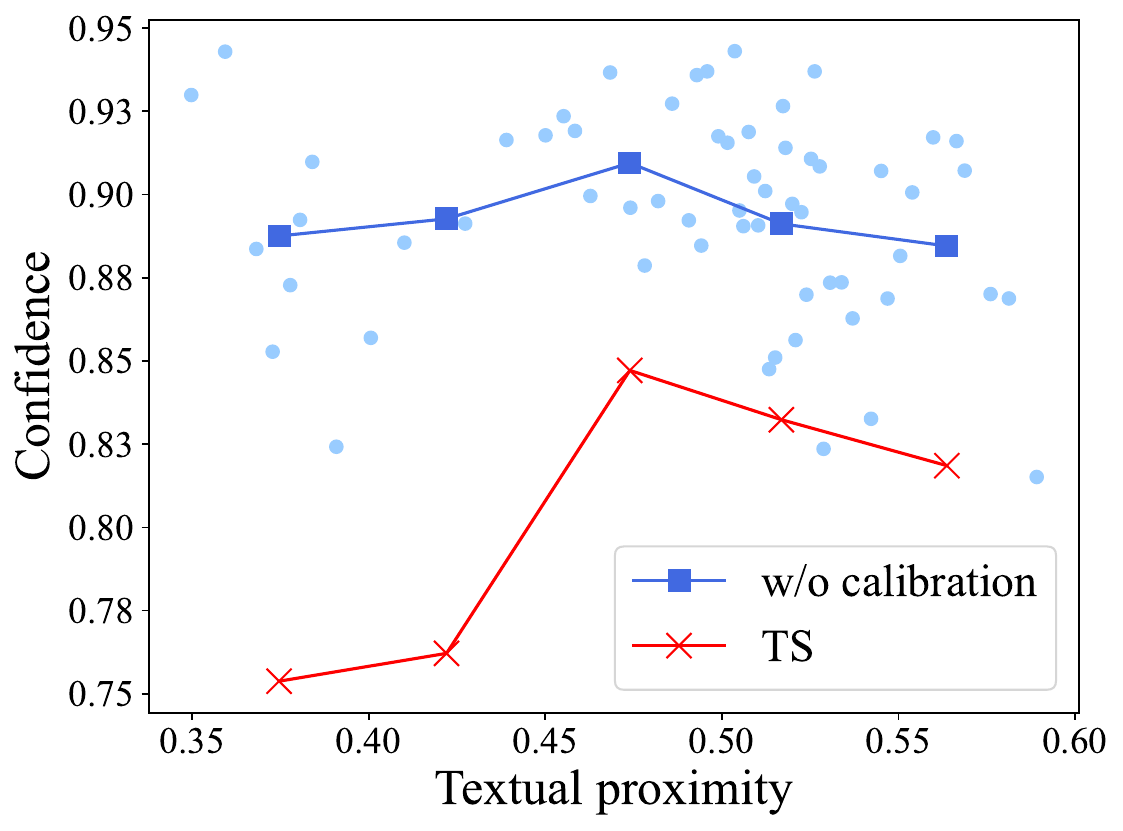}
        \caption{Food101}
    \end{subfigure}
    \hfill
    \begin{subfigure}{.235\textwidth}
        \centering
        \includegraphics[width=\linewidth]{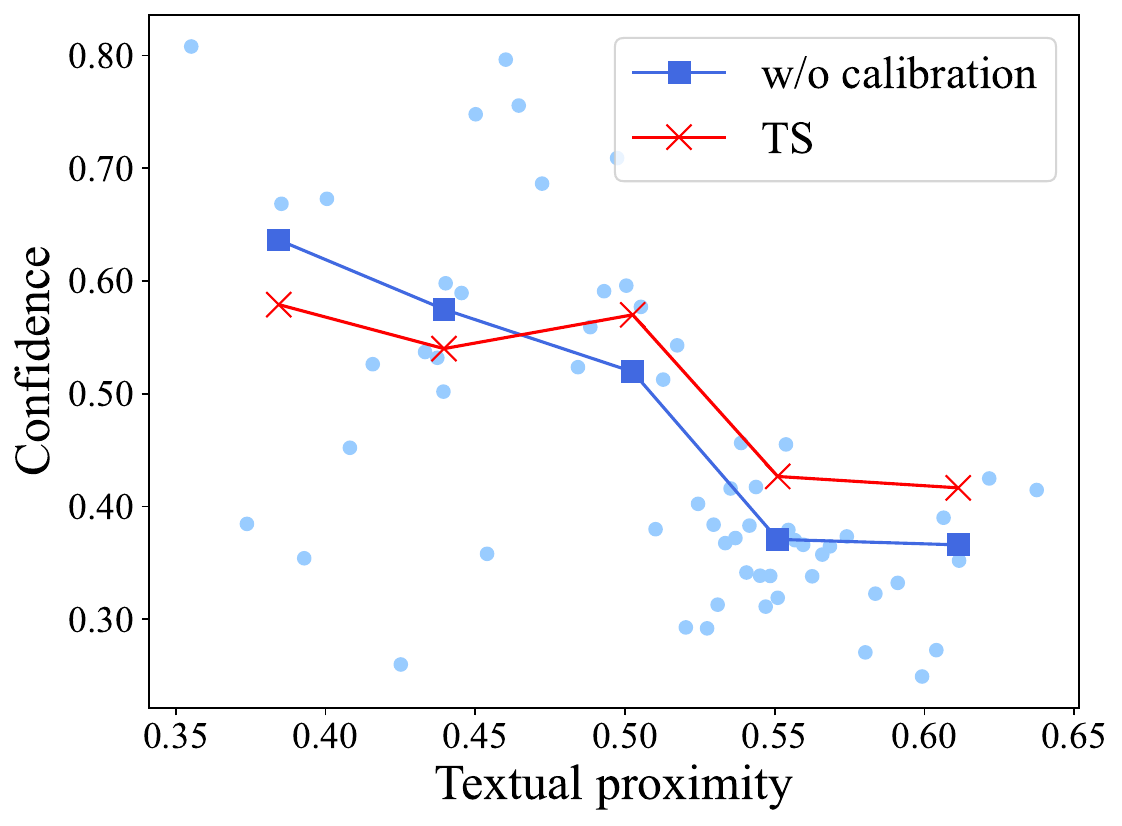}
        \caption{FGVCAircraft}
    \end{subfigure}
    \hfill
    \begin{subfigure}{.235\textwidth}
        \centering
        \includegraphics[width=\linewidth]{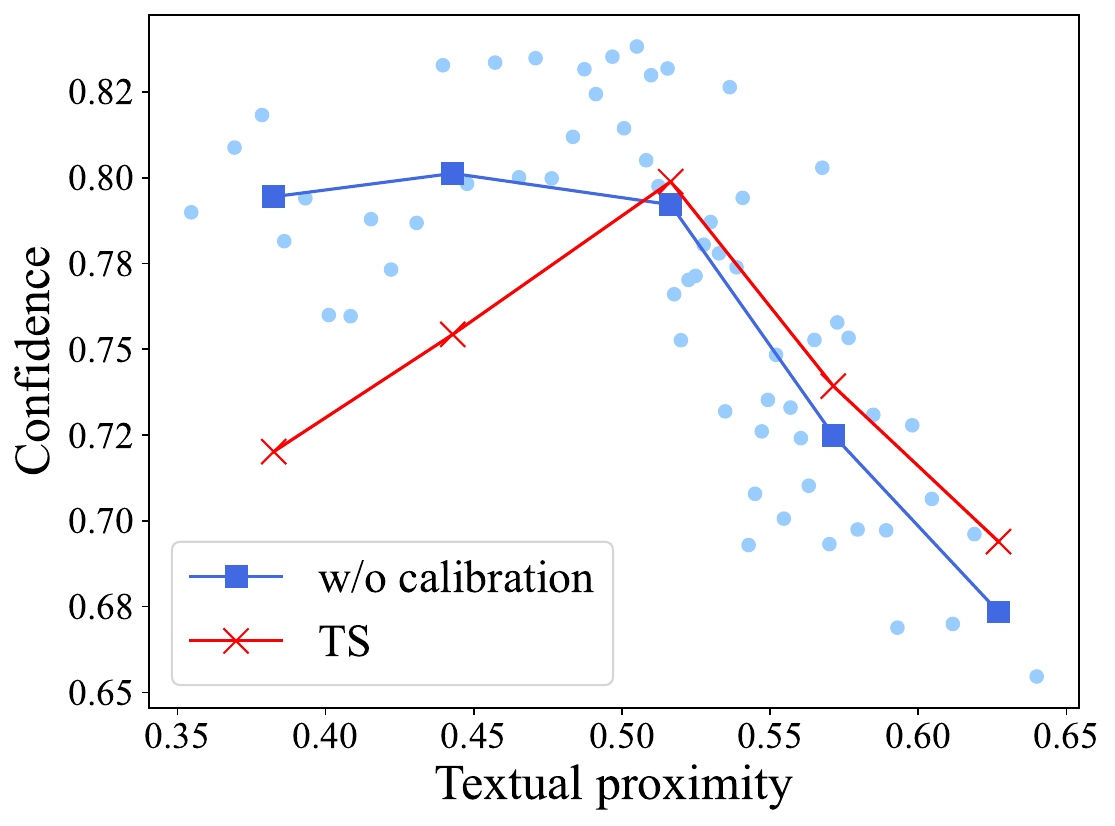}
        \caption{SUN397}
    \end{subfigure}
    \begin{subfigure}{.235\textwidth}
        \centering
        \includegraphics[width=\linewidth]{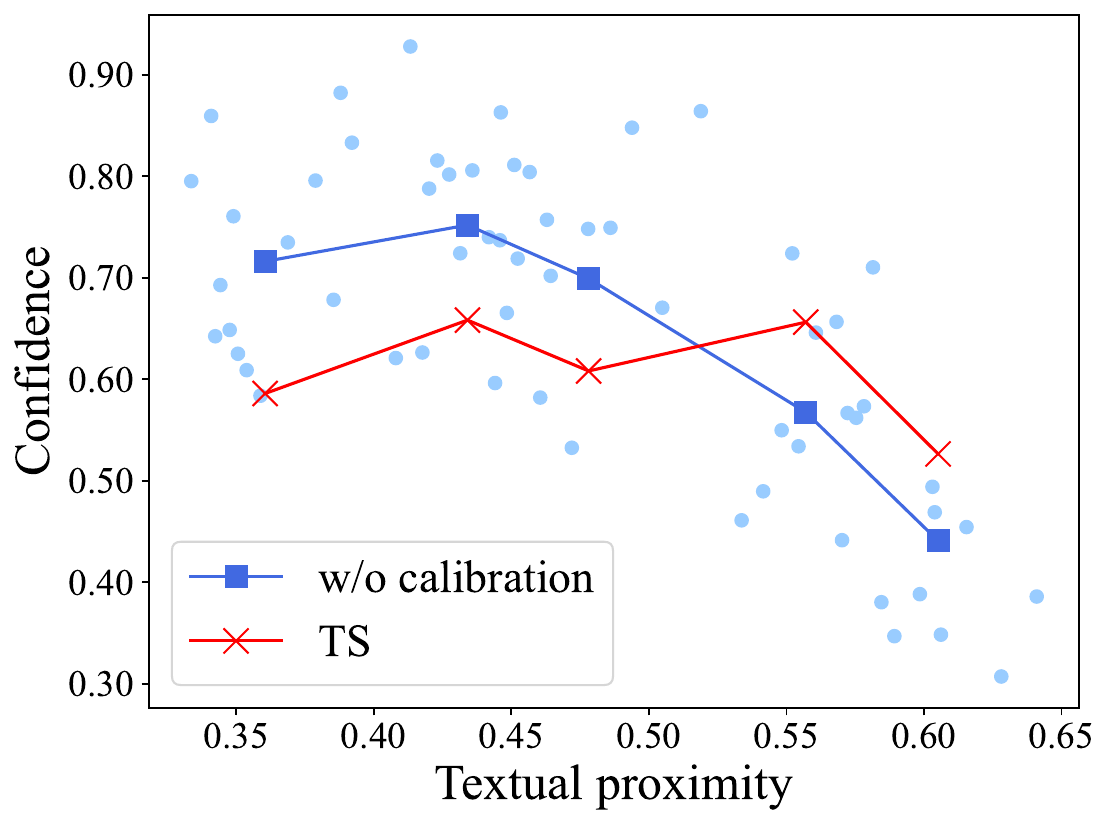}
        \caption{DTD}
    \end{subfigure}
    \hfill
    \begin{subfigure}{.235\textwidth}
        \centering
        \includegraphics[width=\linewidth]{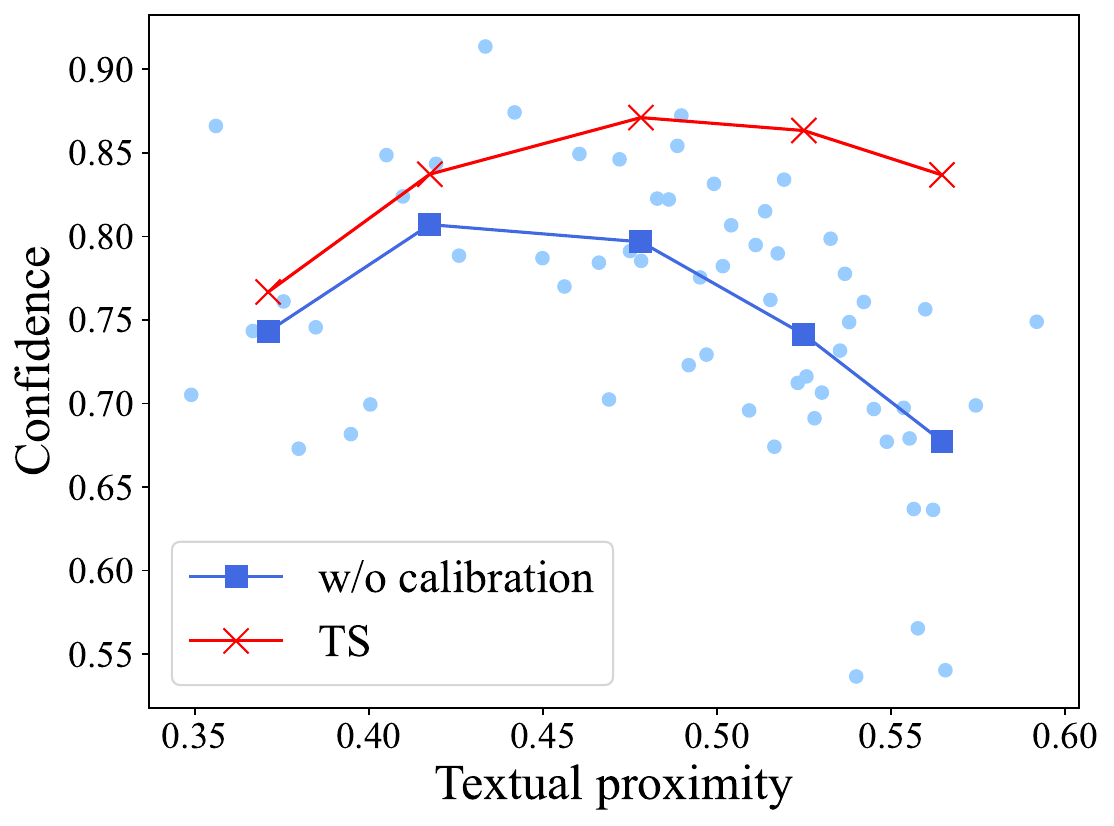}
        \caption{UCF101}
    \end{subfigure}
    \hfill
    \begin{subfigure}{.235\textwidth}
        \centering
        \includegraphics[width=\linewidth]{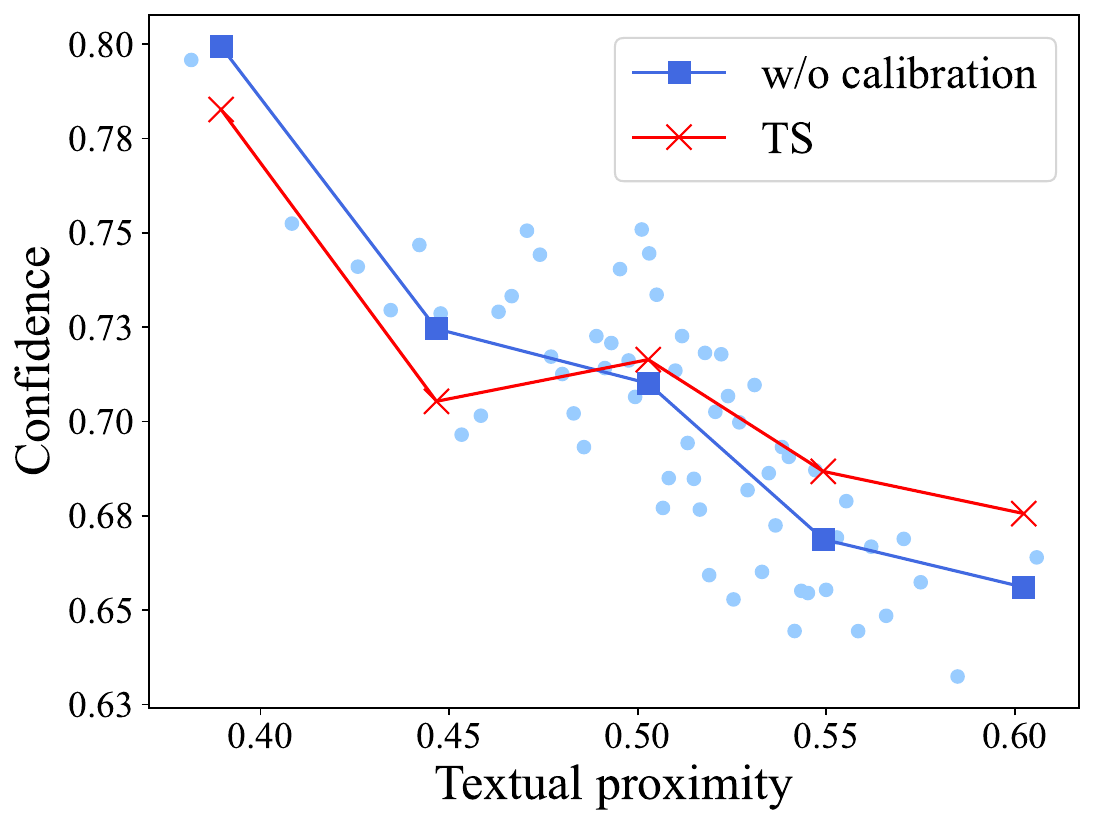}
        \caption{ImageNet}
    \end{subfigure}
    \hfill

    \caption{Class-wise confidence on downstream dataset after tuning.}
    \label{appx_p_confidence}
\end{figure}

\begin{figure}[h]
    \centering
    \begin{subfigure}{.235\textwidth}
        \centering
        \includegraphics[width=\linewidth]{figs/p/p_StanfordCars_ece.pdf}
        \caption{StanfordCars}
    \end{subfigure}
    \hfill
    \begin{subfigure}{.235\textwidth}
        \centering
        \includegraphics[width=\linewidth]{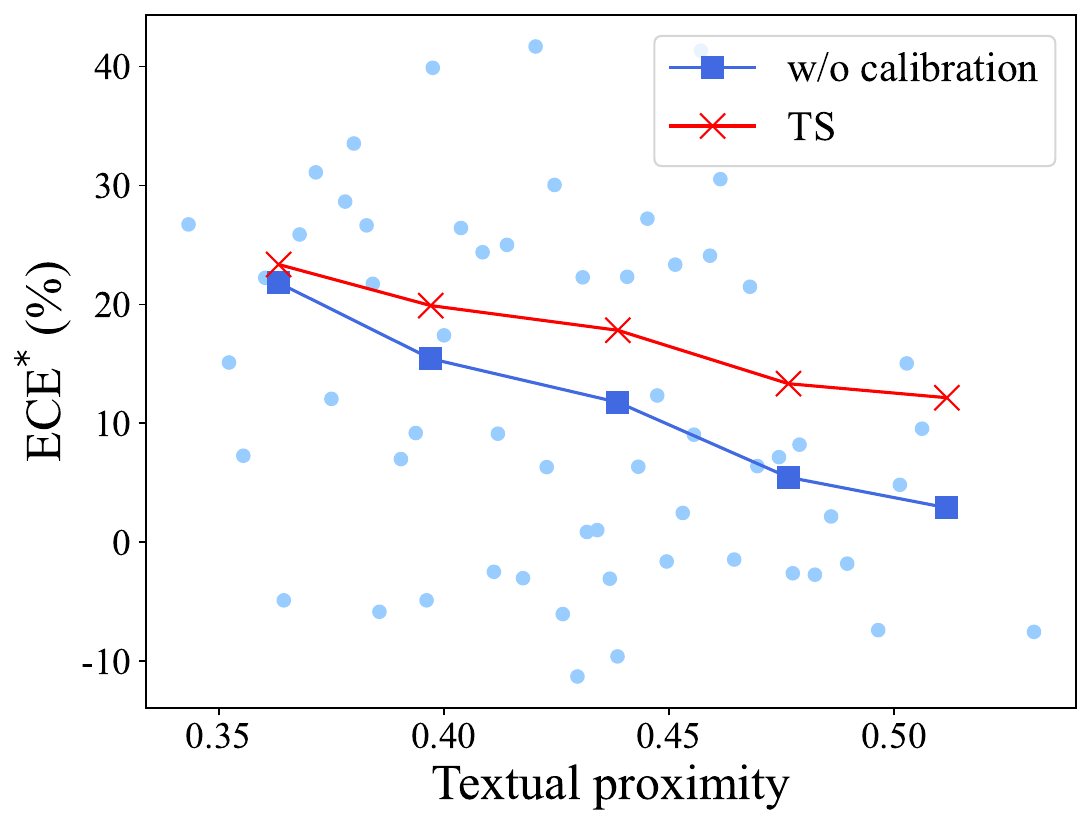}
        \caption{Flowers102}
    \end{subfigure}
    \hfill
    \begin{subfigure}{.235\textwidth}
        \centering
        \includegraphics[width=\linewidth]{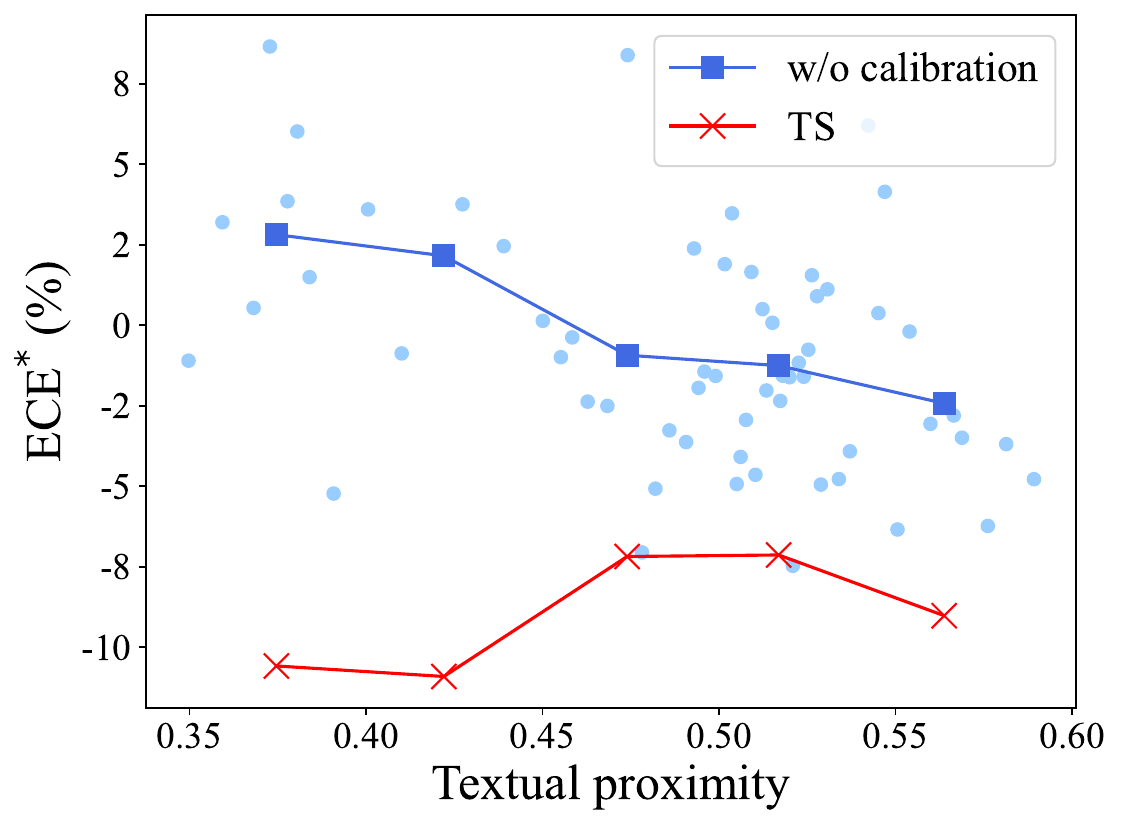}
        \caption{Food101}
    \end{subfigure}
    \hfill
    \begin{subfigure}{.235\textwidth}
        \centering
        \includegraphics[width=\linewidth]{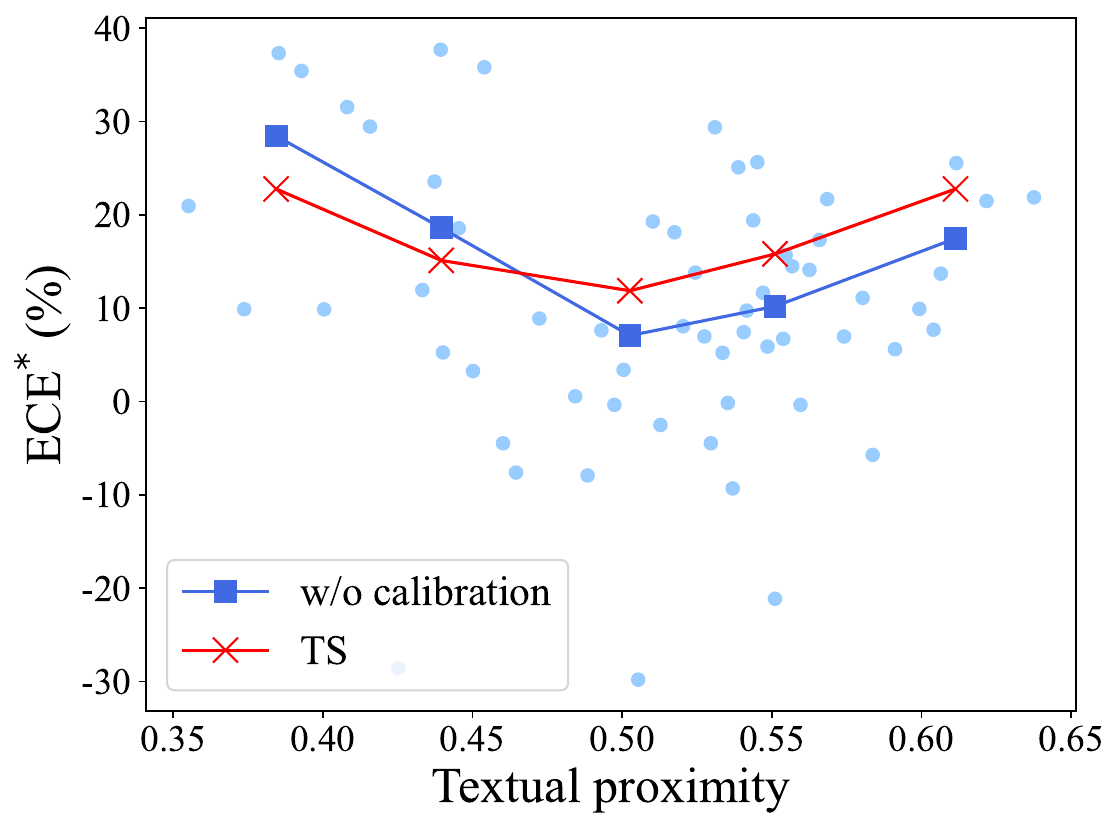}
        \caption{FGVCAircraft}
    \end{subfigure}
    \hfill
    \begin{subfigure}{.235\textwidth}
        \centering
        \includegraphics[width=\linewidth]{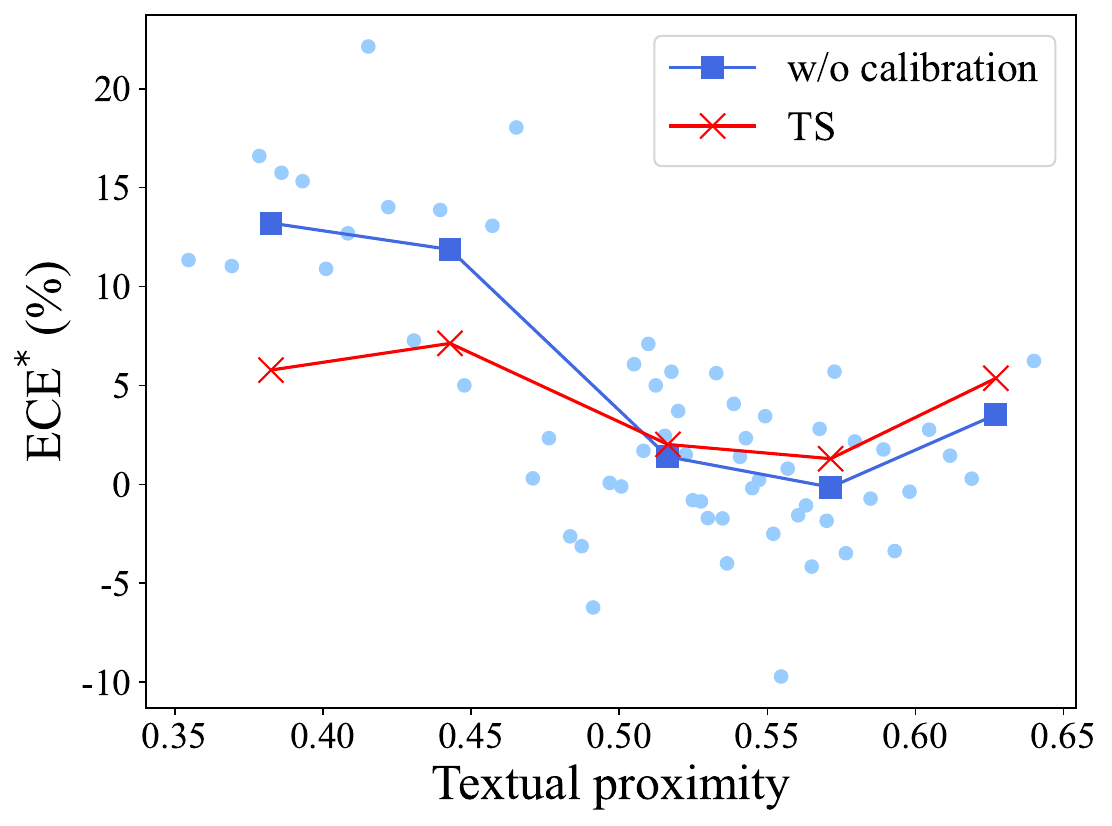}
        \caption{SUN397}
    \end{subfigure}
    \begin{subfigure}{.235\textwidth}
        \centering
        \includegraphics[width=\linewidth]{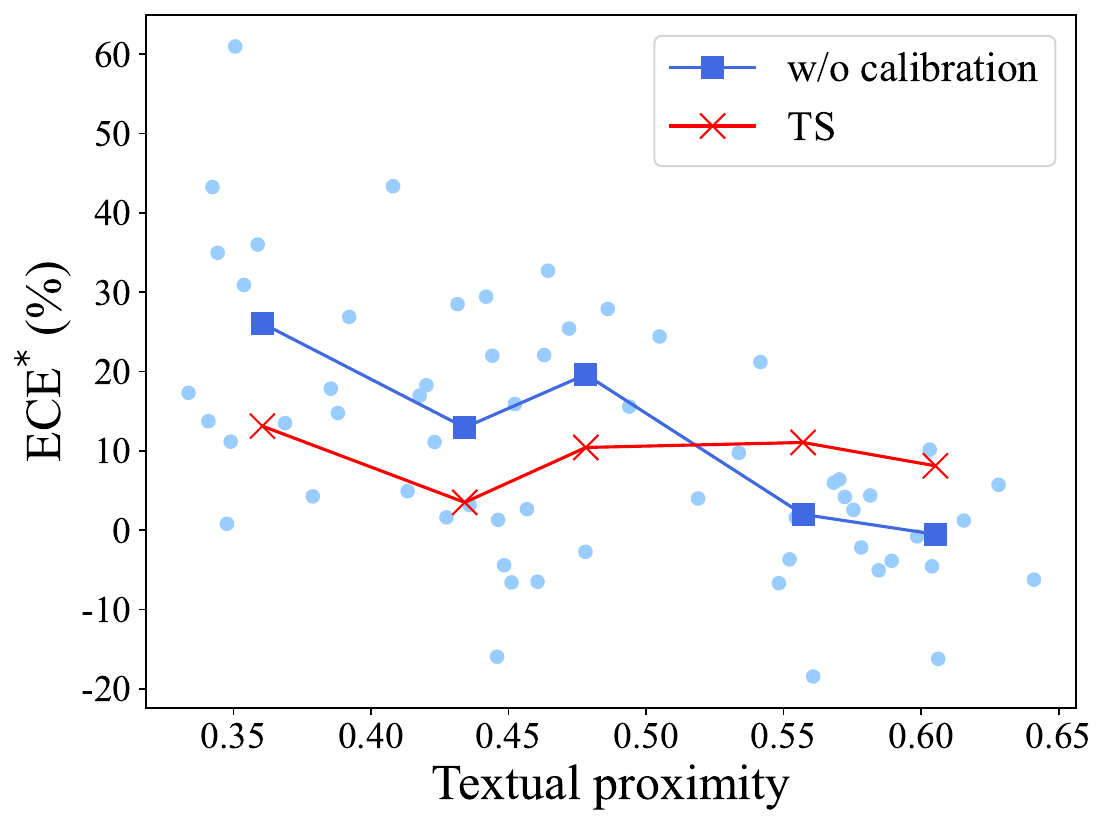}
        \caption{DTD}
    \end{subfigure}
    \hfill
    \begin{subfigure}{.235\textwidth}
        \centering
        \includegraphics[width=\linewidth]{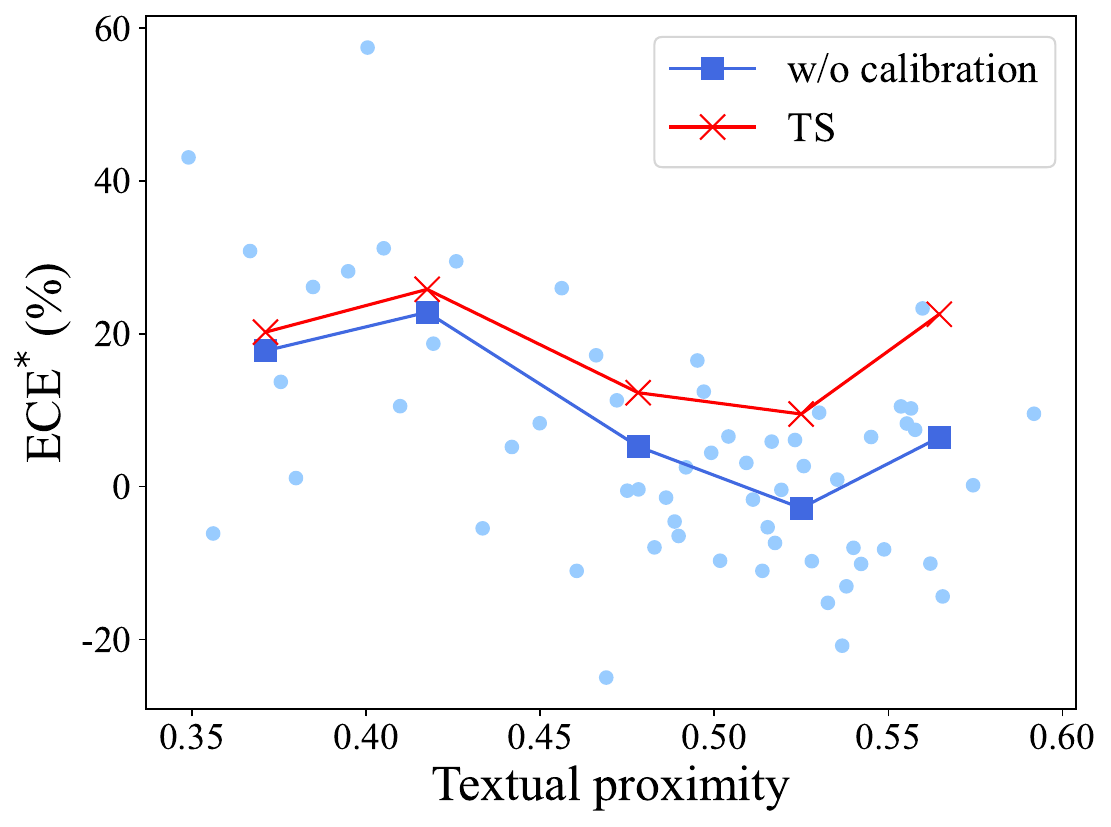}
        \caption{UCF101}
    \end{subfigure}
    \hfill
    \begin{subfigure}{.235\textwidth}
        \centering
        \includegraphics[width=\linewidth]{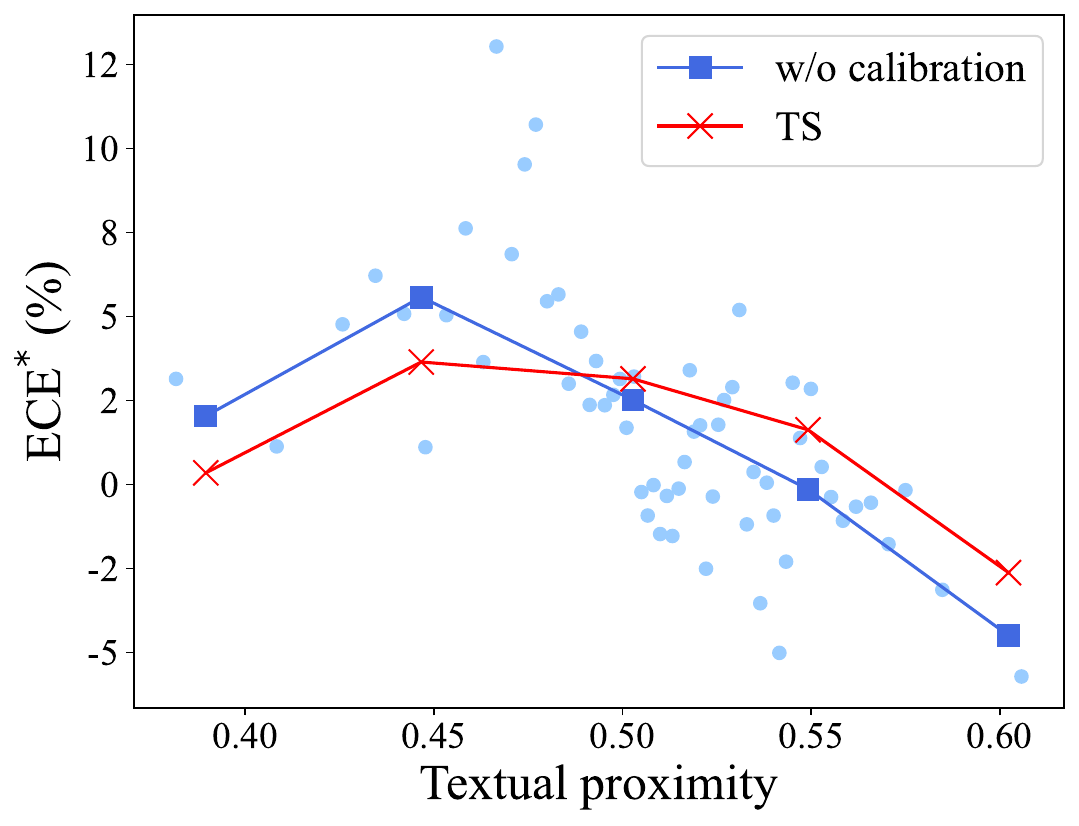}
        \caption{ImageNet}
    \end{subfigure}
    \hfill

    \caption{Class-wise performance on downstream dataset after tuning.}
    \label{appx_p_ece}
\end{figure}

To showcase that the deviation degree in the textual gap is crucial for open-vocabulary calibration, we first introduce a distance-based metric named \textit{proximity} (see Section. \ref{sec_motivation}) to quantify the deviation degree of textual features.
We present the correlation between the deviation degree of textual feature and confidence / ECE in Table \ref{appx_p_confidence} and \ref{appx_p_ece}. Here we use $\text{ECE}^{*}$ with a positive (negative) value denoting overconfidence (underconfidence) for a better view.
The results show that Lower proximity correlates with higher confidence and ECE.

\subsubsection{Confidence}

\clearpage
\subsubsection{$\text{ECE}^{*}$}

\subsection{TD score}

\begin{figure}[h]
    \centering
    \begin{subfigure}{.235\textwidth}
        \centering
        \includegraphics[width=\linewidth]{figs/s/s_StanfordCars_ece.pdf}
        \caption{StanfordCars}
    \end{subfigure}
    \hfill
    \begin{subfigure}{.235\textwidth}
        \centering
        \includegraphics[width=\linewidth]{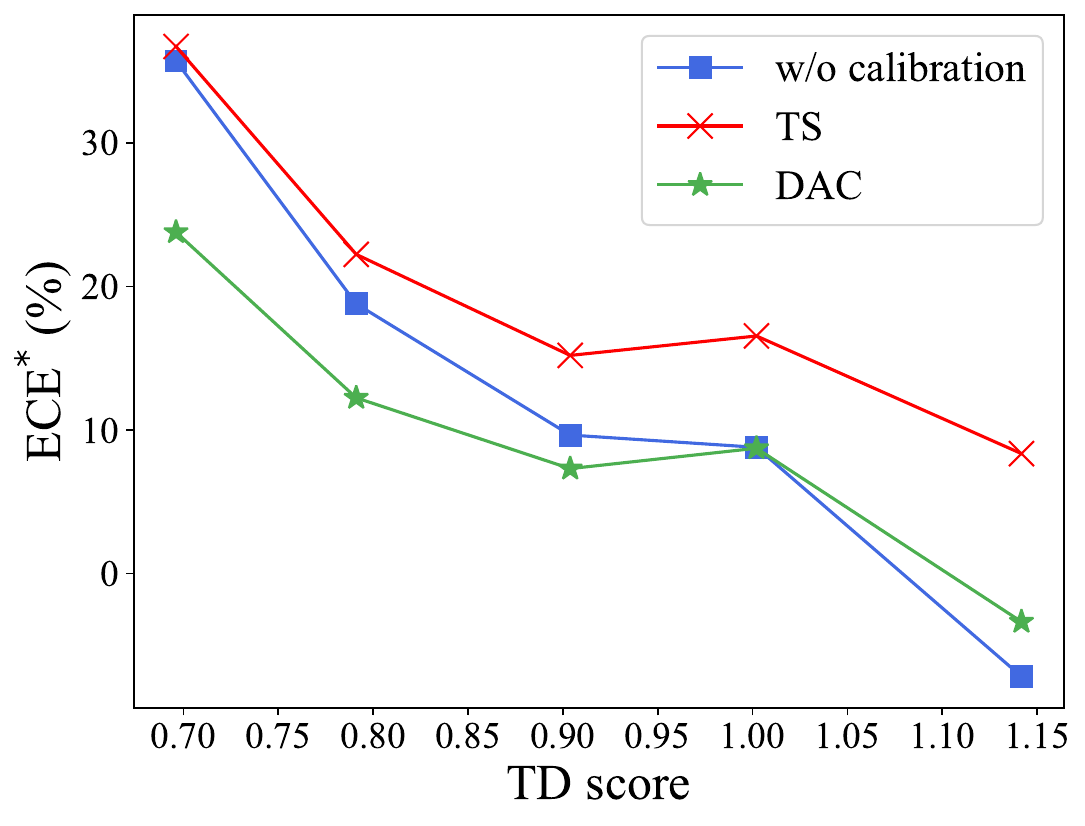}
        \caption{Flowers102}
    \end{subfigure}
    \hfill
    \begin{subfigure}{.235\textwidth}
        \centering
        \includegraphics[width=\linewidth]{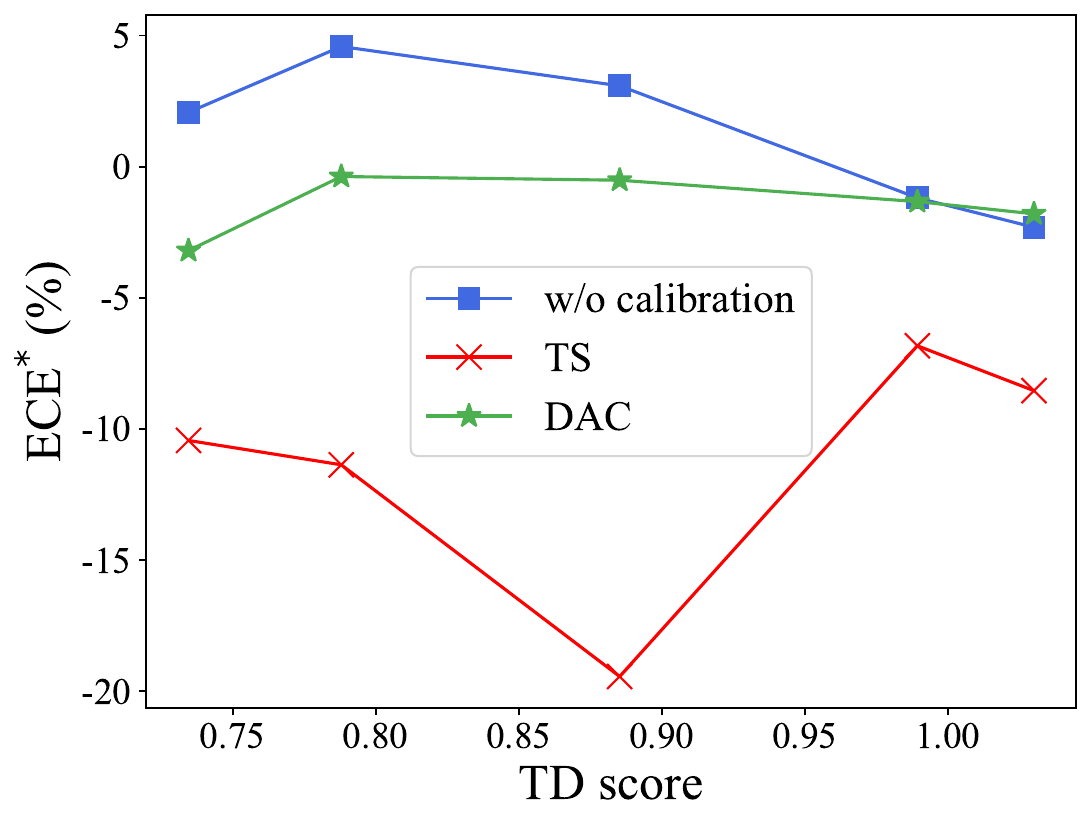}
        \caption{Food101}
    \end{subfigure}
    \hfill
    \begin{subfigure}{.235\textwidth}
        \centering
        \includegraphics[width=\linewidth]{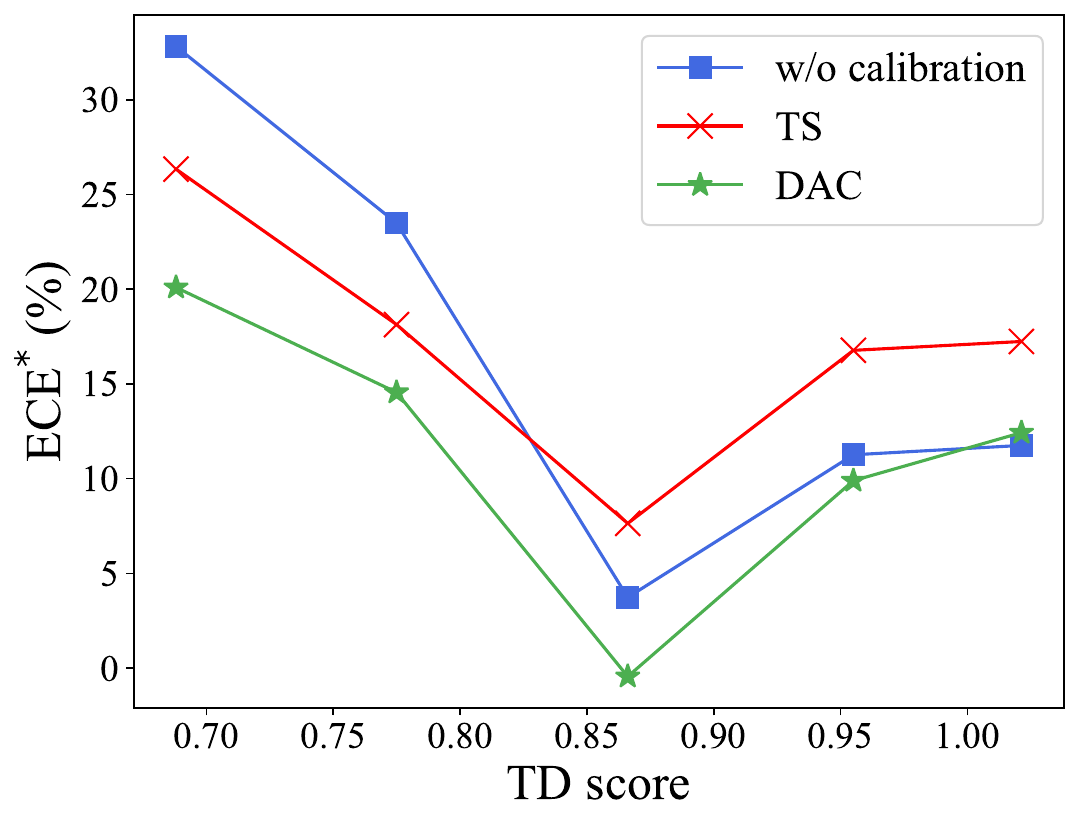}
        \caption{FGVCAircraft}
    \end{subfigure}
    \hfill
    \begin{subfigure}{.235\textwidth}
        \centering
        \includegraphics[width=\linewidth]{figs/s/s_SUN397_ece.pdf}
        \caption{SUN397}
    \end{subfigure}
    \begin{subfigure}{.235\textwidth}
        \centering
        \includegraphics[width=\linewidth]{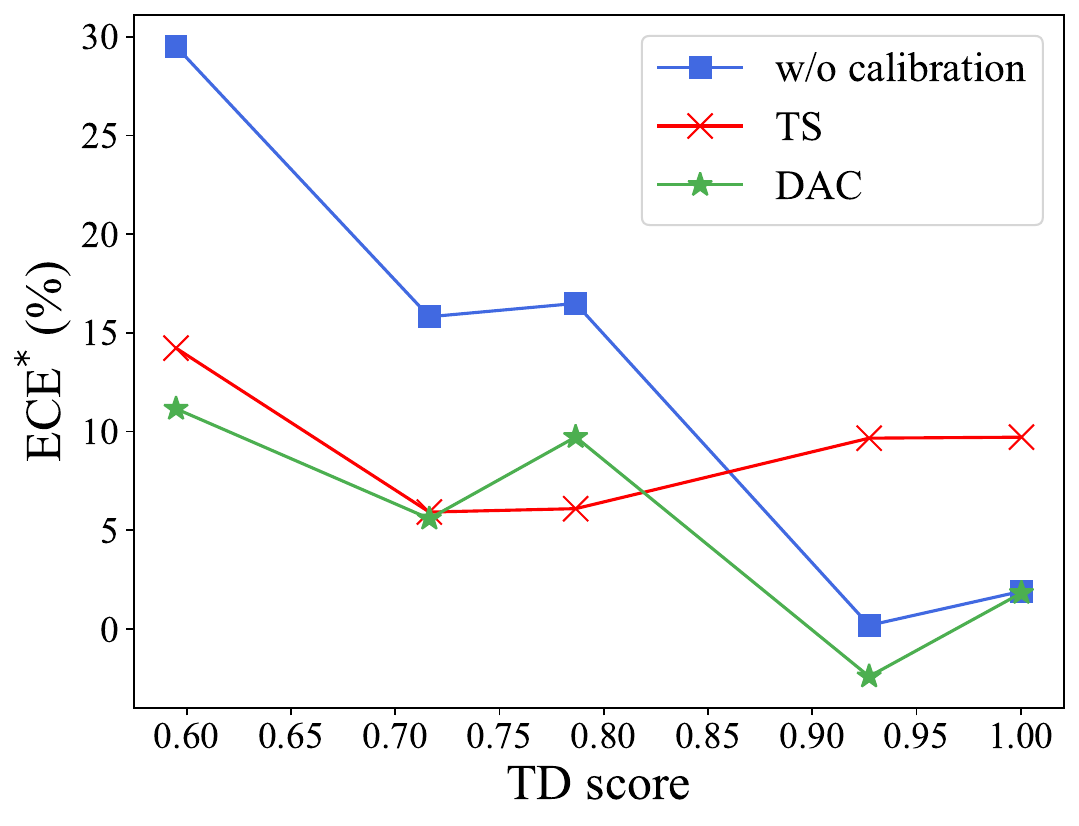}
        \caption{DTD}
    \end{subfigure}
    \hfill
    \begin{subfigure}{.235\textwidth}
        \centering
        \includegraphics[width=\linewidth]{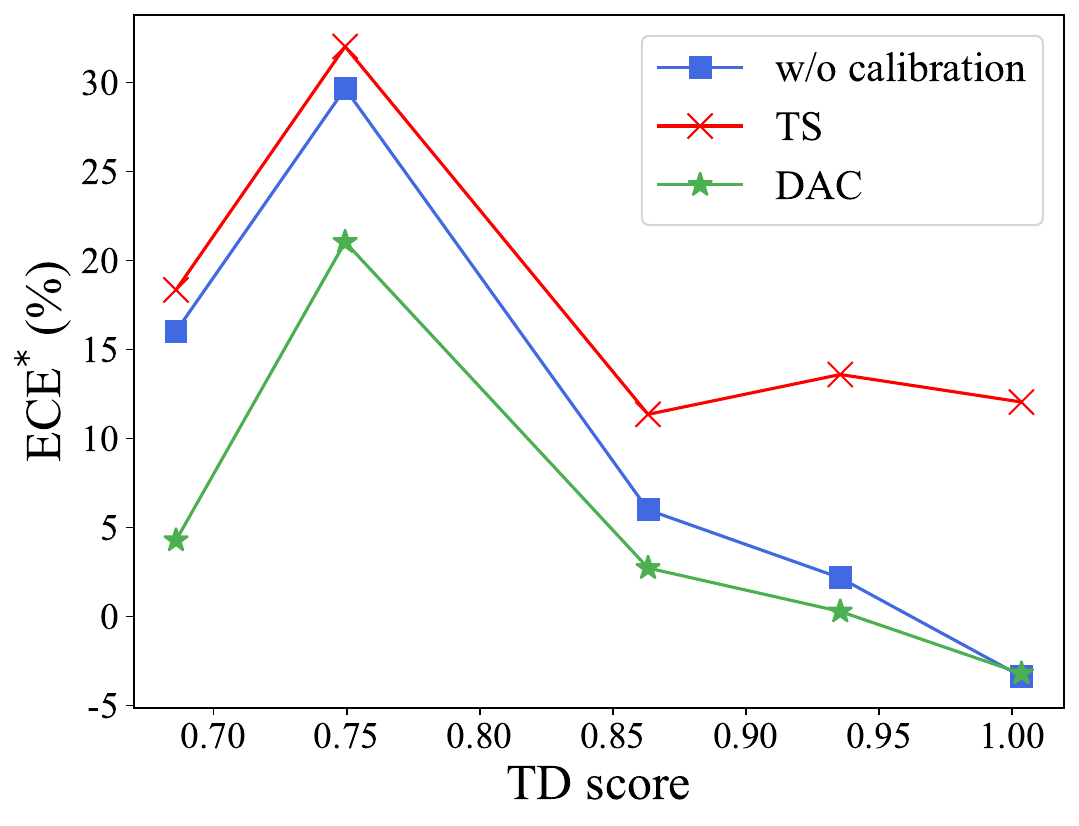}
        \caption{UCF101}
    \end{subfigure}
    \hfill
    \begin{subfigure}{.235\textwidth}
        \centering
        \includegraphics[width=\linewidth]{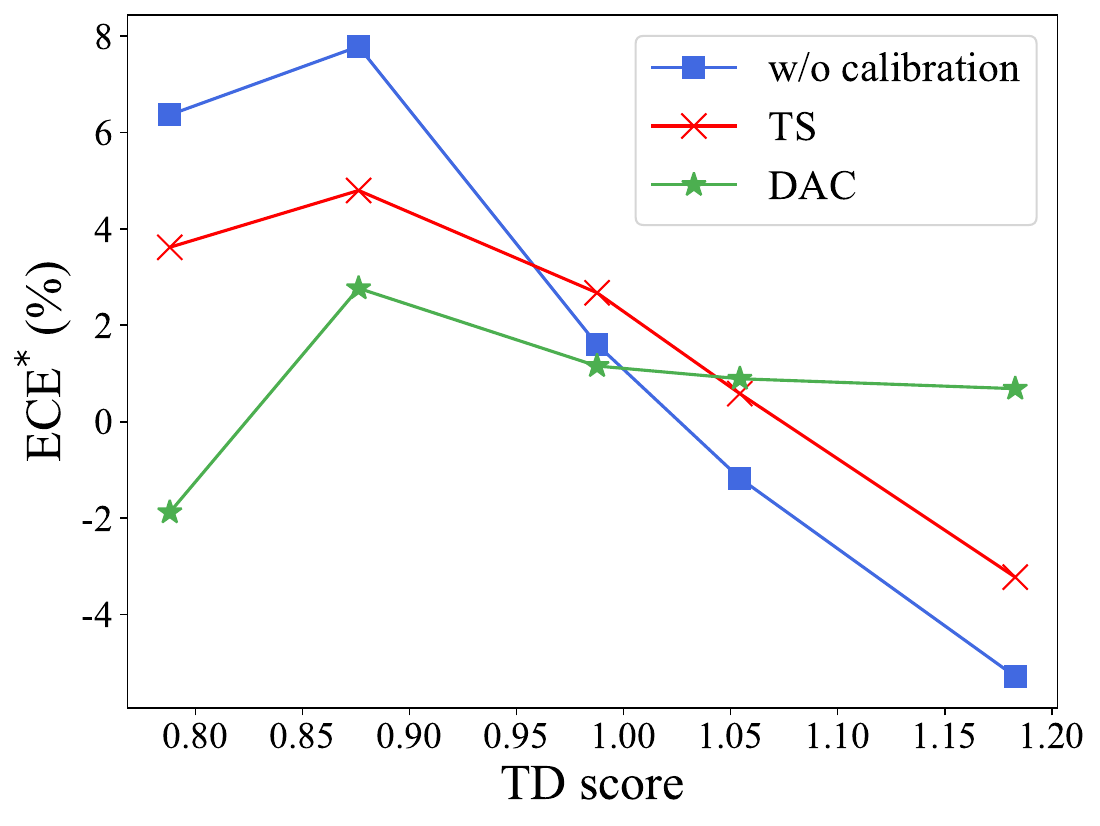}
        \caption{ImageNet}
    \end{subfigure}
    \hfill

    \caption{Class-wise performance on downstream dataset after tuning.}
    \label{appx_td}
\end{figure}

To verify the effectiveness of our proposed method, we present the class-wise results with DAC.
As the TD score decreases in Figure \ref{appx_td}, the tuned model shows higher confidence in open-vocabulary
classes and TS cannot help the issue. In particular, DAC addresses this overconfidence by reducing the logit scale for the prediction with low textual proximity, thereby ensuring more reliable predictions.

\label{sec_td_score}


\clearpage
\section{Detailed Experimental Results}
In this section, we illustrate the experimental results in detail.

\subsection{The calibration performance of tuned VLMs}
\label{calibration_fs}
To showcase the miscalibration in tuned VLM, we fine-tune the pre-trained CLIP with 7 tuning methods on 11 downstream datasets and use ECE as the calibration metric. 
We report the detailed calibration results in this section. The datasets and compared methods are listed in Section \ref{sec_exp_setup}.

As is shown in Table \ref{tab_fs_cab_detail}, for the calibration of the base class, CoOp shows the best calibration results in base classes.  
We can also observe that distillation-based tuning (KgCoOp, ProGrad and PromptSRC) restrains the model's confidence in base classes.
For the calibration of the new class, distillation-based tuning utilizes zero-shot CLIP as the teacher model to improve the generalization ability, which generates relatively reliable predictions in unseen classes as expected. 

More importantly, we observe that tuned VLMs tend to be underconfident in base classes while prone to overconfident in new classes. 
This phenomenon motivates us to further investigate the calibration of fine-tuned VLM.


\begin{table}[h]
    \centering
    \caption{Expected Calibration Error (ECE) on diverse downstream datasets using various tuning methods for CLIP-ViT-B/16. The calibration performance is averaged across three variants. ``base'' / ``new'' denotes, the performance on seen/unseen classes.}
    \label{tab:temps}

    \begin{subtable}{\linewidth}
        \centering
    \resizebox{1.0\textwidth}{!}{
   \begin{tabular}{ccccccccccccc}
    \toprule
    Methods & Caltech101  & OxfordPets & StanfordCars & Flowers102 & Food101 & FGVCAircraft & SUN397 & DTD   & EuroSAT & UCF101 & ImageNet & AVG \\
    \midrule
    ZeroshotCLIP & 6.49  & 2.25  & 3.74  & 3.11  & 1.57  & 3.03  & 1.59  & 4.53  & 8.35  & 3.24  & 1.51  & 3.58  \\
    CoOp  & 0.75  & 0.97  & 2.76  & 2.26  & 2.64  & 5.56  & 4.42  & 6.47  & 1.33  & 3.64  & 1.65  & 2.95  \\
    CoCoOp & 1.72  & 2.29  & 7.01  & 7.41  & 1.02  & 3.66  & 1.84  & 2.85  & 8.21  & 2.17  & 2.58  & 3.71  \\
    ProDA & 1.85  & 3.48  & 11.16  & 5.03  & 1.95  & 8.24  & 3.15  & 2.38  & 9.60  & 4.54  & 4.82  & 5.11  \\
    KgCoOp & 2.32  & 2.96  & 11.36  & 9.97  & 1.35  & 5.88  & 4.68  & 7.97  & 10.84  & 4.17  & 2.63  & 5.83  \\
    MaPLe & 1.12  & 1.92  & 6.65  & 3.66  & 0.74  & 3.06  & 1.22  & 3.12  & 4.35  & 2.29  & 2.19  & 2.76  \\
    ProGrad & 2.76  & 2.71  & 9.62  & 8.92  & 1.39  & 5.68  & 3.68  & 10.99  & 10.07  & 5.04  & 2.49  & 5.76  \\
    PromptSRC & 2.24  & 2.39  & 8.05  & 4.59  & 0.96  & 4.68  & 2.23  & 2.77  & 8.95  & 2.52  & 2.01  & 3.76  \\
    \bottomrule
    \end{tabular}%
        }
    \subcaption{Base}
    \label{tab:week1}

    \end{subtable}


    \begin{subtable}{\linewidth}
        \centering
            \resizebox{1.0\textwidth}{!}{
   \begin{tabular}{ccccccccccccc}
    \toprule
    Methods & Caltech101  & OxfordPets & StanfordCars & Flowers102 & Food101 & FGVCAircraft & SUN397 & DTD   & EuroSAT & UCF101 & ImageNet & AVG \\
    \midrule
    ZeroshotCLIP & 1.60  & 3.42  & 3.31  & 4.91  & 1.83  & 6.55  & 3.48  & 6.86  & 9.12  & 5.52  & 2.09  & 4.43  \\
    CoOp  & 4.08  & 1.83  & 12.50  & 18.34  & 3.83  & 28.44  & 13.70  & 26.86  & 12.71  & 19.24  & 10.69 & 13.84  \\
    CoCoOp & 3.89  & 2.19  & 2.31  & 11.49  & 1.70  & 11.26  & 2.20  & 15.71  & 12.32  & 4.48  & 1.65  & 6.29  \\
    ProDA & 4.61  & 6.60  & 3.02  & 6.67  & 1.51  & 5.36  & 1.49  & 8.22  & 4.84  & 3.11  & 1.59  & 4.27  \\
    KgCoOp & 1.96  & 3.16  & 2.81  & 5.99  & 1.98  & 12.53  & 1.28  & 7.03  & 6.82  & 2.64  & 1.8   & 4.36  \\
    MaPLe & 1.33  & 2.51  & 2.02  & 13.43  & 1.26  & 10.38  & 3.01  & 15.71  & 9.06  & 2.38  & 2.39  & 5.77  \\
    ProGrad & 1.91  & 2.98  & 2.88  & 5.45  & 2.04  & 13.18  & 1.75  & 4.22  & 7.40  & 2.85  & 1.81  & 4.22  \\
    PromptSRC & 1.62  & 3.08  & 2.16  & 5.77  & 1.60  & 9.79  & 0.77  & 5.79  & 7.35  & 2.64  & 1.71  & 3.84  \\
    \bottomrule
    \end{tabular}%
        }
    \subcaption{New}
    \end{subtable}
\end{table}
    \label{tab_fs_cab_detail}
\newpage

\subsection{Calibrated performance for tuned VLM}
\label{appx_fs_cab}
We utilize commonly used post-hoc confidence calibration for tuned VLMs and expect to calibrate them well on both base and new classes. We can observe that established calibration can remedy miscalibration in base classes. However, such efficacy can not transfer to new classes.

\vspace{-0.5em}
\subsubsection{base class}

\vspace{-0.5em}

\begin{table}[!ht]
    \centering
    \caption{Expected Calibration Error (ECE) on base classes in diverse downstream datasets using various tuning methods for CLIP-ViT-B/16, with calibration performance averaged across three variants.}

    \begin{subtable}{\linewidth}
    \centering
    \resizebox{0.9\textwidth}{!}{
    \begin{tabular}{ccccccccccccc}
    \toprule
    Methods & Caltech101  & OxfordPets & StanfordCars & Flowers102 & Food101 & FGVCAircraft & SUN397 & DTD   & EuroSAT & UCF101 & ImageNet & AVG \\
    \midrule
    ZeroshotCLIP & 6.49  & 2.25  & 3.74  & 3.11  & 1.57  & 3.03  & 1.59  & 4.53  & 8.35  & 3.24  & 1.51  & 3.58  \\
    CoOp  & 0.75  & 0.97  & 2.76  & 2.26  & 2.64  & 5.56  & 4.42  & 6.47  & 1.33  & 3.64  & 1.65  & 2.95  \\
    CoCoOp & 1.72  & 2.29  & 7.01  & 7.41  & 1.02  & 3.66  & 1.84  & 2.85  & 8.21  & 2.17  & 2.58  & 3.71  \\
    ProDA & 1.85  & 3.48  & 11.16  & 5.03  & 1.95  & 8.24  & 3.15  & 2.38  & 9.60  & 4.54  & 4.82  & 5.11  \\
    KgCoOp & 2.32  & 2.96  & 11.36  & 9.97  & 1.35  & 5.88  & 4.68  & 7.97  & 10.84  & 4.17  & 2.63  & 5.83  \\
    MaPLe & 1.12  & 1.92  & 6.65  & 3.66  & 0.74  & 3.06  & 1.22  & 3.12  & 4.35  & 2.29  & 2.19  & 2.76  \\
    ProGrad & 2.76  & 2.71  & 9.62  & 8.92  & 1.39  & 5.68  & 3.68  & 10.99  & 10.07  & 5.04  & 2.49  & 5.76  \\
    PromptSRC & 2.24  & 2.39  & 8.05  & 4.59  & 0.96  & 4.68  & 2.23  & 2.77  & 8.95  & 2.52  & 2.01  & 3.76  \\
    \bottomrule
    \end{tabular}%
    }
    \subcaption{w/ calibation}
    \end{subtable}

    \begin{subtable}{\linewidth}
    \centering
    \resizebox{0.9\textwidth}{!}{
    \begin{tabular}{ccccccccccccc}
    \toprule
    Methods & Caltech101  & OxfordPets & StanfordCars & Flowers102 & Food101 & FGVCAircraft & SUN397 & DTD   & EuroSAT & UCF101 & ImageNet & AVG \\
    \midrule
    ZeroshotCLIP & 1.33  & 0.82  & 1.82  & 2.73  & 8.29  & 2.62  & 1.69  & 4.54  & 7.84  & 7.86  & 1.79  & 3.76  \\
    CoOp  & 0.66  & 1.12  & 1.71  & 0.99  & 4.81  & 2.83  & 1.32  & 5.72  & 1.80  & 5.83  & 1.5   & 2.57  \\
    CoCoOp & 0.96  & 1.14  & 1.45  & 1.15  & 4.00  & 3.07  & 1.37  & 4.74  & 10.90  & 5.43  & 1.56  & 3.25  \\
    ProDA & 0.79  & 1.61  & 2.78  & 1.43  & 4.76  & 3.51  & 1.97  & 2.96  & 6.20  & 5.56  & 1.94  & 3.05  \\
    KgCoOp & 0.62  & 0.67  & 2.26  & 1.87  & 3.76  & 2.62  & 1.65  & 3.88  & 6.43  & 5.60  & 1.48  & 2.80  \\
    MaPLe & 0.51  & 1.15  & 1.89  & 1.77  & 4.40  & 2.71  & 1.21  & 5.98  & 1.93  & 5.22  & 1.91  & 2.61  \\
    ProGrad & 0.93  & 1.18  & 2.07  & 1.30  & 3.26  & 2.64  & 1.21  & 3.41  & 3.34  & 6.47  & 1.23  & 2.46  \\
    PromptSRC & 0.75  & 1.1   & 2.04  & 1.15  & 4.63  & 2.86  & 1.5   & 2.9   & 3.82  & 5.72  & 1.30  & 2.52  \\
    \bottomrule
    \end{tabular}%
    }
    \subcaption{Temperature Scaling}
    \end{subtable}

    \begin{subtable}{\linewidth}
        \centering
    \resizebox{0.9\textwidth}{!}{
    \begin{tabular}{ccccccccccccc}
    \toprule
    Methods & Caltech101  & OxfordPets & StanfordCars & Flowers102 & Food101 & FGVCAircraft & SUN397 & DTD   & EuroSAT & UCF101 & ImageNet & AVG \\
    \midrule
    ZeroshotCLIP & 2.72  & 3.60  & 2.43  & 4.44  & 2.28  & 3.37  & 1.04  & 2.72  & 7.40  & 5.88  & 0.44  & 3.30  \\
    CoOp  & 1.09  & 5.77  & 2.20  & 1.01  & 3.18  & 3.85  & 1.59  & 1.09  & 1.50  & 6.67  & 1.19  & 2.65  \\
    CoCoOp & 0.96  & 1.25  & 2.86  & 1.85  & 2.18  & 4.81  & 1.19  & 0.96  & 5.64  & 4.34  & 0.67  & 2.43  \\
    ProDA & 0.70  & 2.13  & 2.63  & 1.40  & 1.74  & 5.58  & 0.98  & 0.70  & 3.58  & 5.56  & 0.73  & 2.34  \\
    KgCoOp & 0.93  & 1.61  & 2.94  & 2.40  & 1.80  & 5.42  & 1.40  & 0.93  & 3.29  & 4.66  & 0.58  & 2.36  \\
    MaPLe & 0.85  & 3.17  & 1.61  & 1.22  & 1.69  & 4.56  & 1.59  & 0.85  & 5.73  & 3.68  & 0.66  & 2.33  \\
    ProGrad & 0.74  & 1.98  & 2.97  & 0.98  & 1.34  & 5.40  & 1.42  & 0.74  & 26.48  & 5.29  & 0.72  & 4.37  \\
    PromptSRC & 0.84  & 0.65  & 2.05  & 0.79  & 1.77  & 4.27  & 1.17  & 0.84  & 8.71  & 4.27  & 0.80  & 2.38  \\
    \bottomrule
    \end{tabular}%
    }
    \subcaption{Density-Ratio Calibration}
    \end{subtable}

    \begin{subtable}{\linewidth}
        \centering
    \resizebox{0.9\textwidth}{!}{
    \begin{tabular}{ccccccccccccc}
    \toprule
    Methods & Caltech101  & OxfordPets & StanfordCars & Flowers102 & Food101 & FGVCAircraft & SUN397 & DTD   & EuroSAT & UCF101 & ImageNet & AVG \\
    \midrule
    ZeroshotCLIP & 2.20  & 1.54  & 8.81  & 8.87  & 2.87  & 1.70  & 5.18  & 10.93  & 9.27  & 7.37  & 4.78  & 5.77  \\
    CoOp  & 0.89  & 1.39  & 7.92  & 1.21  & 3.42  & 11.44  & 5.26  & 7.79  & 1.35  & 7.21  & 3.95  & 4.71  \\
    CoCoOp & 0.99  & 0.79  & 9.94  & 2.30  & 2.08  & 7.80  & 6.07  & 7.07  & 3.39  & 7.89  & 4.7   & 4.82  \\
    ProDA & 0.87  & 1.46  & 8.74  & 1.98  & 2.28  & 7.27  & 4.81  & 7.54  & 3.23  & 8.03  & 4.23  & 4.59  \\
    KgCoOp & 0.72  & 0.51  & 9.03  & 1.74  & 2.45  & 7.27  & 5.46  & 8.33  & 2.79  & 8.66  & 4.95  & 4.72  \\
    MaPLe & 0.82  & 1.03  & 10.76  & 1.39  & 2.64  & 8.68  & 5.21  & 4.47  & 2.16  & 8.14  & 5.02  & 4.57  \\
    ProGrad & 1.00  & 1.36  & 7.14  & 1.84  & 3.15  & 10.09  & 5.32  & 7.87  & 2.96  & 8.35  & 4.98  & 4.91  \\
    PromptSRC & 0.52  & 1.10  & 7.48  & 1.59  & 2.22  & 10.28  & 4.38  & 5.26  & 3.77  & 5.83  & 4.73  & 4.29  \\
    \bottomrule
    \end{tabular}%
    }
    \subcaption{Histogram Binning}
    \end{subtable}

    \begin{subtable}{\linewidth}
        \centering
    \resizebox{0.9\textwidth}{!}{
    \begin{tabular}{ccccccccccccc}
    \toprule
    Methods & Caltech101  & OxfordPets & StanfordCars & Flowers102 & Food101 & FGVCAircraft & SUN397 & DTD   & EuroSAT & UCF101 & ImageNet & AVG \\
    \midrule
    ZeroshotCLIP & 3.07  & 6.97  & 9.28  & 4.99  & 6.46  & 8.96  & 7.76  & 7.34  & 8.56  & 11.71  & 2.08  & 7.02  \\
    CoOp  & 5.23  & 4.70  & 11.85  & 2.00  & 8.46  & 12.23  & 9.36  & 8.33  & 8.17  & 15.68  & 1.16  & 7.92  \\
    CoCoOp & 5.05  & 4.96  & 9.45  & 3.70  & 10.17  & 12.20  & 8.14  & 9.70  & 9.93  & 13.77  & 1.81  & 8.08  \\
    ProDA & 3.98  & 4.20  & 8.69  & 3.03  & 8.26  & 9.98  & 7.26  & 6.48  & 11.88  & 16.61  & 2.09  & 7.50  \\
    KgCoOp & 1.32  & 5.37  & 9.15  & 2.72  & 8.91  & 10.88  & 7.57  & 6.52  & 9.53  & 15.45  & 1.79  & 7.20  \\
    MaPLe & 5.43  & 5.63  & 11.03  & 1.49  & 9.29  & 11.54  & 8.22  & 5.07  & 6.29  & 16.70  & 1.49  & 7.47  \\
    ProGrad & 2.47  & 5.50  & 8.72  & 3.10  & 9.72  & 12.41  & 7.25  & 8.17  & 11.34  & 16.88  & 1.77  & 7.94  \\
    PromptSRC & 3.56  & 4.23  & 10.09  & 1.86  & 10.11  & 12.40  & 8.18  & 7.28  & 12.76  & 10.25  & 1.53  & 7.48  \\
    \bottomrule
    \end{tabular}%
    }
    \subcaption{Isotonic Regression}
    \end{subtable}

    \begin{subtable}{\linewidth}
        \centering
    \resizebox{0.9\textwidth}{!}{
    \begin{tabular}{ccccccccccccc}
    \toprule
    Methods & Caltech101  & OxfordPets & StanfordCars & Flowers102 & Food101 & FGVCAircraft & SUN397 & DTD   & EuroSAT & UCF101 & ImageNet & AVG \\
    \midrule
    ZeroshotCLIP & 3.54  & 1.07  & 3.19  & 2.43  & 2.81  & 1.97  & 2.03  & 3.91  & 2.86  & 4.26  & 0.31  & 2.58  \\
    CoOp  & 0.54  & 1.23  & 2.69  & 1.10  & 1.36  & 3.60  & 2.17  & 2.54  & 1.05  & 5.09  & 0.19  & 1.96  \\
    CoCoOp & 0.71  & 0.78  & 2.71  & 2.36  & 1.94  & 3.26  & 2.49  & 4.74  & 6.61  & 3.32  & 0.48  & 2.67  \\
    ProDA & 0.69  & 0.79  & 4.13  & 1.54  & 2.22  & 5.42  & 1.67  & 3.89  & 5.39  & 3.88  & 0.82  & 2.77  \\
    KgCoOp & 0.84  & 1.13  & 4.60  & 2.20  & 1.40  & 3.35  & 2.13  & 4.49  & 3.34  & 3.59  & 0.48  & 2.50  \\
    MaPLe & 0.55  & 1.72  & 2.97  & 1.36  & 1.64  & 4.19  & 1.62  & 4.44  & 1.40  & 2.95  & 0.33  & 2.11  \\
    ProGrad & 0.81  & 1.98  & 3.47  & 1.56  & 1.77  & 3.33  & 2.10  & 3.45  & 1.41  & 4.03  & 0.39  & 2.21  \\
    PromptSRC & 0.91  & 0.85  & 3.20  & 0.82  & 1.29  & 4.54  & 1.76  & 3.11  & 5.06  & 2.74  & 0.27  & 2.23  \\
    \bottomrule
    \end{tabular}%
    }
    \subcaption{Multi-Isotonic Regression}
    \end{subtable}

\end{table}

\paragraph{Post-hoc calibration can remedy miscalibration in base classes.} 
In the latest works \cite{tu2023closer, oh2023towards}, scaling-based methods like TS have been successfully adopted in VLM calibration, and their findings reveal that this calibration can handle the miscalibration well in the close-world setting.
As a complement, we observe that both scaling-based (e.g., DEN) and bin-based (e.g., MIR) calibration can notably reduce miscalibration in base classes. 

\paragraph{Post-hoc calibration on base classes can not transfer to new classes.} 
For instance, scaling-based methods like TS scale the logits by a single scalar temperature value. If tuned VLMs show underconfidence in base classes, TS sharpens the logit scale before the softmax function and makes the prediction more confident.
As a cost, the confidence distribution of tuned VLM becomes sharper and causes the VLMs to display exaggerated overconfidence in new classes.
for bin-based methods, they need the probabilities from base classes as the input, which are incompatible with zero-shot prediction.

\subsubsection{new class}
\label{tab_dac}

\begin{table}[H]
    \centering
    \caption{Expected Calibration Error (ECE) on new classes in diverse downstream datasets using various tuning methods for CLIP-ViT-B/16, with calibration performance averaged across three variants.}
    \begin{subtable}{\linewidth}
        \centering
            \resizebox{0.95\textwidth}{!}{
   \begin{tabular}{ccccccccccccc}
    \toprule
    Methods & Caltech101  & OxfordPets & StanfordCars & Flowers102 & Food101 & FGVCAircraft & SUN397 & DTD   & EuroSAT & UCF101 & ImageNet & AVG \\
    \midrule
    ZeroshotCLIP & 1.60  & 3.42  & 3.31  & 4.91  & 1.83  & 6.55  & 3.48  & 6.86  & 9.12  & 5.52  & 2.09  & 4.43  \\
    CoOp  & 4.08  & 1.83  & 12.50  & 18.34  & 3.83  & 28.44  & 13.70  & 26.86  & 12.71  & 19.24  & 10.69 & 13.84  \\
    CoCoOp & 3.89  & 2.19  & 2.31  & 11.49  & 1.70  & 11.26  & 2.20  & 15.71  & 12.32  & 4.48  & 1.65  & 6.29  \\
    ProDA & 4.61  & 6.60  & 3.02  & 6.67  & 1.51  & 5.36  & 1.49  & 8.22  & 4.84  & 3.11  & 1.59  & 4.27  \\
    KgCoOp & 1.96  & 3.16  & 2.81  & 5.99  & 1.98  & 12.53  & 1.28  & 7.03  & 6.82  & 2.64  & 1.8   & 4.36  \\
    MaPLe & 1.33  & 2.51  & 2.02  & 13.43  & 1.26  & 10.38  & 3.01  & 15.71  & 9.06  & 2.38  & 2.39  & 5.77  \\
    ProGrad & 1.91  & 2.98  & 2.88  & 5.45  & 2.04  & 13.18  & 1.75  & 4.22  & 7.40  & 2.85  & 1.81  & 4.22  \\
    PromptSRC & 1.62  & 3.08  & 2.16  & 5.77  & 1.60  & 9.79  & 0.77  & 5.79  & 7.35  & 2.64  & 1.71  & 3.84  \\
    \bottomrule
    \end{tabular}%
        }
    \subcaption{w/o calibration}
    \end{subtable}
    
    \begin{subtable}{\linewidth}
        \centering
    \resizebox{0.95\textwidth}{!}{
    \begin{tabular}{ccccccccccccc}
    \toprule
    Methods & Caltech101  & OxfordPets & StanfordCars & Flowers102 & Food101 & FGVCAircraft & SUN397 & DTD   & EuroSAT & UCF101 & ImageNet & AVG \\
    \midrule
    ZeroshotCLIP & 3.00  & 1.54  & 1.68  & 5.14  & 8.29  & 7.60  & 3.34  & 10.86  & 11.07  & 1.62  & 3.1   & 5.20  \\
    PromptSRC & 2.6   & 1.87  & 6.72  & 13.45 & 4.55  & 12.91 & 1.3   & 5.8   & 10.05 & 12.72 & 2.48  & 6.77  \\
    ProGrad & 2.66  & 1.48  & 10.14  & 18.50  & 3.16  & 21.31  & 4.10  & 14.75  & 16.70  & 16.54  & 3.43  & 10.25  \\
    ProDA & 5.66  & 3.99  & 7.26  & 14.38  & 5.12  & 15.17  & 1.11  & 6.49  & 6.42  & 15.20  & 3.9   & 7.70  \\
    MaPLe & 1.61  & 1.86  & 4.89  & 18.02  & 3.89  & 13.15  & 2.12  & 6.68  & 11.04  & 7.45  & 2.76  & 6.68  \\
    KgCoOp & 3.04  & 1.36  & 5.72  & 16.21  & 3.54  & 21.33  & 2.55  & 14.24  & 10.42  & 12.37  & 2.57  & 8.49  \\
    CoOp  & 4.43  & 2.28  & 15.78  & 25.16  & 6.01  & 24.21  & 7.44  & 10.67  & 15.71  & 23.47  & 7.33  & 12.95  \\
    CoCoOp & 4.55  & 1.11  & 6.38  & 19.95  & 3.88  & 17.50  & 2.13  & 12.36  & 19.01  & 11.78  & 2.32  & 9.18  \\
    \bottomrule
    \end{tabular}%
    }
    \subcaption{Temperature Scaling}
    \end{subtable}

    \begin{subtable}{\linewidth}
        \centering
    \resizebox{0.95\textwidth}{!}{
    \begin{tabular}{ccccccccccccc}
    \toprule
    Methods & Caltech101  & OxfordPets & StanfordCars & Flowers102 & Food101 & FGVCAircraft & SUN397 & DTD   & EuroSAT & UCF101 & ImageNet & AVG \\
    \midrule
    ZeroshotCLIP & 3.61  & 1.13  & 4.01  & 7.23  & 2.62  & 6.07  & 3.54  & 10.57  & 9.02  & 5.16  & 2.19  & 5.01  \\
    PromptSRC & 3.86  & 1.42  & 6.17  & 12.96  & 1.68  & 16.34  & 2.70  & 10.35  & 24.47  & 9.42  & 3.04  & 8.40  \\
    ProGrad & 3.90  & 0.96  & 8.36  & 18.32  & 1.40  & 22.28  & 4.62  & 14.08  & 69.06  & 13.13  & 3.77  & 14.53  \\
    ProDA & 6.40  & 2.82  & 6.12  & 14.70  & 2.73  & 12.93  & 2.72  & 10.51  & 13.39  & 11.67  & 3.86  & 7.99  \\
    MaPLe & 2.01  & 1.82  & 4.86  & 18.13  & 1.41  & 13.96  & 4.16  & 16.19  & 26.05  & 4.93  & 4.22  & 8.89  \\
    KgCoOp & 4.43  & 1.20  & 5.88  & 14.48  & 1.54  & 21.33  & 3.94  & 16.05  & 16.16  & 10.86  & 2.76  & 8.97  \\
    CoOp  & 5.89  & 4.95  & 16.35  & 24.31  & 4.81  & 26.77  & 9.57  & 23.31  & 16.90 & 26.56  & 8.08  & 15.23  \\
    CoCoOp & 5.32  & 1.62  & 5.33  & 19.03  & 2.02  & 17.18  & 4.35  & 16.11  & 9.47  & 9.75  & 2.18  & 8.40  \\
    \bottomrule
    \end{tabular}%
    }
    \subcaption{Density-Ratio Calibration}
    \end{subtable}
    \begin{subtable}{\linewidth}
        \centering
    \resizebox{0.95\textwidth}{!}{
    \begin{tabular}{ccccccccccccc}
    \toprule
    Methods & Caltech101  & OxfordPets & StanfordCars & Flowers102 & Food101 & FGVCAircraft & SUN397 & DTD   & EuroSAT & UCF101 & ImageNet & AVG \\
    \midrule
    ZeroshotCLIP & 1.60  & 3.42  & 3.31  & 4.91  & 1.83  & 6.55  & 3.48  & 6.86  & 9.12  & 5.52  & 2.09  & 4.43  \\
    CoOp  & 3.17  & 1.82  & 5.16  & 10.19  & 1.78  & 17.38  & 4.05  & 10.48  & 8.62  & 8.67  & 5.67  & 7.00  \\
    CoCoOp & 3.65  & 2.43  & 2.21  & 7.74  & 1.64  & 9.03  & 1.09  & 7.47  & 13.49  & 2.70  & 1.61  & 4.82  \\
    ProDA & 4.87  & 4.72  & 3.28  & 6.32  & 0.70  & 7.40  & 1.06  & 5.68  & 3.33  & 4.14  & 2.42  & 3.99  \\
    KgCoOp & 1.84  & 3.11  & 3.12  & 5.90  & 1.94  & 11.78  & 1.67  & 7.09  & 6.59  & 2.69  & 1.83  & 4.32  \\
    MaPLe & 1.26  & 2.51  & 2.75  & 11.28  & 1.50  & 9.06  & 1.22  & 8.16  & 8.55  & 2.30  & 2.11  & 4.61  \\
    ProGrad & 1.97  & 3.31  & 2.29  & 5.04  & 1.85  & 10.46  & 1.32  & 3.49  & 6.90  & 2.42  & 2.11  & 3.74  \\
    PromptSRC & 1.58  & 2.98  & 2.39  & 5.03  & 1.55  & 8.55  & 0.79  & 5.50  & 7.24  & 2.46  & 1.84  & 3.63  \\
    \bottomrule
    \end{tabular}%
    }
    \subcaption{Distance-Aware Calibration}
    \end{subtable}
    
\end{table}

\newpage
\subsection{Accuracy \& Confidence of tuned VLMs}
\label{fs_acc_conf}

In this section, we analyze the miscalibration from the classification performance. 
Since open-vocabulary classes are not explicitly seen during the tuning, some might believe that this miscalibration arises from rapid accuracy decreasing and leads to a larger Expected Calibration Error (ECE). 
This hypothesis leads to an open question: "\textit{Should we blame accuracy degeneration for miscalibration?}" 

We report the average performance on 11 datasets in Table \ref{tab_accuracy} and \ref{tab_conf}. Although the classification performance on new classes may be comparable or lower, the average predictive confidence of tuned CLIP is still relatively higher than zero-shot CLIP, which results in a larger ECE. Moreover, we take a close look at the confidence level of MaPLe in new classes. Although the accuracy is comparable with zero-shot CLIP,  we can find that the confidence level is greatly improved, which indicates the overconfidence issue.


\subsubsection{Accuracy} 

\begin{table}[h]
    \centering
    \caption{Accuracy comparison of existing prompt tuning in the base-to-new generalization setting.}
    \begin{subtable}{\linewidth}
        \centering
    \resizebox{0.95\textwidth}{!}{
    \begin{tabular}{ccccccccccccc}
    \toprule
    Methods & Caltech101  & OxfordPets & StanfordCars & Flowers102 & Food101 & FGVCAircraft & SUN397 & DTD   & EuroSAT & UCF101 & ImageNet & AVG \\
    \midrule
    ZeroshotCLIP & 97.20  & 91.30  & 63.60  & 71.80  & 90.10  & 27.70  & 69.40  & 53.00  & 57.00  & 71.00  & 72.40  & 69.50  \\
    CoOp  & 98.23  & 93.57  & 80.07  & 97.90  & 87.83  & 42.57  & 80.60  & 80.80  & 92.63  & 84.77  & 75.60  & 83.14  \\
    CoCoOp & 97.73  & 95.07  & 70.77  & 94.73  & 90.57  & 35.77  & 79.47  & 77.00  & 87.27  & 82.40  & 76.03  & 80.62  \\
    ProDA & 98.53  & 96.13  & 76.43  & 97.93  & 90.77  & 39.77  & 82.37  & 82.80  & 90.47  & 85.43  & 77.57  & 83.47  \\
    KgCoOp & 97.73  & 94.93  & 75.33  & 96.40  & 90.57  & 38.53  & 80.87  & 81.23  & 89.57  & 84.33  & 76.00  & 82.32  \\
    MaPLe & 97.93  & 95.27  & 72.80  & 96.43  & 90.60  & 35.50  & 80.93  & 80.87  & 93.57  & 82.90  & 77.00  & 82.16  \\
    ProGrad & 98.30  & 94.47  & 78.77  & 96.23  & 90.37  & 40.43  & 81.33  & 76.43  & 90.40  & 84.70  & 76.80  & 82.57  \\
    PromptSRC & 98.27  & 95.57  & 80.30  & 98.27  & 90.63  & 44.37  & 83.10  & 83.80  & 92.70  & 87.73  & 77.83  & 84.78  \\
    \bottomrule
    \end{tabular}%
    }
    \subcaption{Base}
    
    \end{subtable}

    \begin{subtable}{\linewidth}
        \centering
    \resizebox{0.95\textwidth}{!}{
    \begin{tabular}{ccccccccccccc}
    \toprule
    Methods & Caltech101  & OxfordPets & StanfordCars & Flowers102 & Food101 & FGVCAircraft & SUN397 & DTD   & EuroSAT & UCF101 & ImageNet & AVG \\
    \midrule
    ZeroshotCLIP & 94.10  & 97.10  & 75.00  & 77.50  & 91.10  & 35.90  & 75.50  & 60.60  & 63.80  & 78.60  & 68.10  & 74.30  \\
    CoOp  & 91.00  & 91.33  & 58.53  & 57.10  & 84.30  & 22.93  & 63.70  & 44.37  & 65.30  & 57.30  & 59.07  & 63.18  \\
    CoCoOp & 93.13  & 97.70  & 72.70  & 69.80  & 91.33  & 33.47  & 76.60  & 53.63  & 62.37  & 74.23  & 70.63  & 72.33  \\
    ProDA & 92.47  & 97.37  & 74.23  & 75.90  & 89.10  & 34.87  & 78.00  & 58.60  & 66.57  & 72.60  & 71.40  & 73.74  \\
    KgCoOp & 94.20  & 97.40  & 74.17  & 72.83  & 91.60  & 30.40  & 75.33  & 50.10  & 66.67  & 74.70  & 69.73  & 72.47  \\
    MaPLe & 95.30  & 97.53  & 75.07  & 71.50  & 91.63  & 36.20  & 77.10  & 57.83  & 67.40  & 78.67  & 70.60  & 74.44  \\
    ProGrad & 93.73  & 97.47  & 69.40  & 71.80  & 90.10  & 29.37  & 72.40  & 51.57  & 60.57  & 71.83  & 67.33  & 70.51  \\
    PromptSRC & 94.03  & 97.37  & 75.33  & 76.57  & 91.47  & 37.53  & 78.63  & 61.83  & 70.73  & 78.13  & 70.23  & 75.62  \\
    \bottomrule
    \end{tabular}%
    }
    \subcaption{New}
    \end{subtable}
    \label{tab_accuracy}
\end{table}

\subsubsection{Confidence}
\begin{table}[H]
    \centering
    \caption{Confidence comparison of existing prompt tuning in the base-to-new generalization setting.}
    \begin{subtable}{\linewidth}
        \centering
    \resizebox{0.95\textwidth}{!}{
    \begin{tabular}{ccccccccccccc}
    \toprule
    Methods & Caltech101  & OxfordPets & StanfordCars & Flowers102 & Food101 & FGVCAircraft & SUN397 & DTD   & EuroSAT & UCF101 & ImageNet & AVG \\
    \midrule
    ZeroshotCLIP & 0.91  & 0.89  & 0.60  & 0.73  & 0.88  & 0.27  & 0.70  & 0.52  & 0.64  & 0.72  & 0.72  & 0.69  \\
    CoOp  & 0.98  & 0.93  & 0.77  & 0.96  & 0.90  & 0.48  & 0.85  & 0.87  & 0.91  & 0.89  & 0.77  & 0.85  \\
    CoCoOp & 0.97  & 0.93  & 0.64  & 0.88  & 0.90  & 0.32  & 0.78  & 0.76  & 0.79  & 0.81  & 0.73  & 0.77  \\
    ProDA & 0.97  & 0.93  & 0.65  & 0.93  & 0.89  & 0.32  & 0.79  & 0.82  & 0.81  & 0.82  & 0.73  & 0.79  \\
    KgCoOp & 0.96  & 0.92  & 0.64  & 0.86  & 0.89  & 0.33  & 0.76  & 0.73  & 0.79  & 0.80  & 0.73  & 0.76  \\
    MaPLe & 0.97  & 0.94  & 0.66  & 0.93  & 0.90  & 0.35  & 0.81  & 0.83  & 0.89  & 0.83  & 0.75  & 0.81  \\
    ProGrad & 0.96  & 0.92  & 0.69  & 0.87  & 0.89  & 0.35  & 0.78  & 0.65  & 0.80  & 0.80  & 0.74  & 0.77  \\
    PromptSRC & 0.96  & 0.93  & 0.72  & 0.94  & 0.90  & 0.40  & 0.81  & 0.82  & 0.84  & 0.86  & 0.76  & 0.81  \\
    \bottomrule
    \end{tabular}%
    }
    \subcaption{Base}
    \end{subtable}

    \begin{subtable}{\linewidth}
        \centering
    \resizebox{0.95\textwidth}{!}{
    \begin{tabular}{ccccccccccccc}
    \toprule
    Methods & Caltech101  & OxfordPets & StanfordCars & Flowers102 & Food101 & FGVCAircraft & SUN397 & DTD   & EuroSAT & UCF101 & ImageNet & AVG \\
    \midrule
    ZeroshotCLIP & 0.95  & 0.94  & 0.72  & 0.79  & 0.89  & 0.41  & 0.72  & 0.52  & 0.62  & 0.73  & 0.68  & 0.72  \\
    CoOp  & 0.95  & 0.92  & 0.71  & 0.75  & 0.88  & 0.51  & 0.78  & 0.71  & 0.75  & 0.77  & 0.70  & 0.77  \\
    CoCoOp & 0.96  & 0.95  & 0.74  & 0.81  & 0.90  & 0.45  & 0.78  & 0.70  & 0.71  & 0.78  & 0.70  & 0.77  \\
    ProDA & 0.97  & 0.91  & 0.74  & 0.83  & 0.88  & 0.40  & 0.77  & 0.66  & 0.63  & 0.74  & 0.71  & 0.75  \\
    KgCoOp & 0.95  & 0.94  & 0.72  & 0.78  & 0.90  & 0.43  & 0.74  & 0.56  & 0.69  & 0.74  & 0.69  & 0.74  \\
    MaPLe & 0.95  & 0.95  & 0.75  & 0.85  & 0.90  & 0.46  & 0.80  & 0.73  & 0.69  & 0.80  & 0.73  & 0.78  \\
    ProGrad & 0.94  & 0.95  & 0.72  & 0.77  & 0.88  & 0.42  & 0.74  & 0.53  & 0.68  & 0.73  & 0.69  & 0.73  \\
    PromptSRC & 0.95  & 0.94  & 0.76  & 0.82  & 0.90  & 0.47  & 0.78  & 0.66  & 0.69  & 0.80  & 0.71  & 0.77 \\  
    \bottomrule
    \end{tabular}%
    }
    \subcaption{New}
    \end{subtable}
    \label{tab_conf} 
\end{table}

\newpage
\subsection{Detailed results of main experiment} 
\label{appx_main_exp}
In this section, We present the detailed results on the calibration of new classes to verify that DAC can open-vocabulary calibration in existing prompt tuning. 
For comprehensive evaluation, 4 standard metrics are used in our evaluation of open-vocabulary confidence calibration: Expected Calibration Error (ECE) \cite{guo2017calibration},  Maximum Calibration Error (MCE) \cite{guo2017calibration}, Adaptive Calibration Error (ACE) \cite{nixon2019measuring} and Proximity-Informed Expected Calibration Error \cite{xiong2023proximity}.

\begin{table}[H]
    \centering
    \caption{Expected Calibration Error (ECE) on new classes in downstream datasets using various tuning methods for CLIP-ViT-B/16.}
    \begin{subtable}{\linewidth}
        \centering
    \resizebox{1.0\textwidth}{!}{
    \begin{tabular}{ccccccccccccc}
    \toprule
    Methods & Caltech101  & OxfordPets & StanfordCars & Flowers102 & Food101 & FGVCAircraft & SUN397 & DTD   & EuroSAT & UCF101 & ImageNet & AVG \\
    \midrule
    ZeroshotCLIP & 1.60  & 3.42  & 3.31  & 4.91  & 1.83  & 6.55  & 3.48  & 6.86  & 9.12  & 5.52  & 2.09  & 4.43  \\
    CoOp  & 4.08  & 1.83  & 12.50  & 18.34  & 3.83  & 28.44  & 13.70  & 26.86  & 12.71  & 19.24  & 10.69 & 13.84  \\
    CoCoOp & 3.89  & 2.19  & 2.31  & 11.49  & 1.70  & 11.26  & 2.20  & 15.71  & 12.32  & 4.48  & 1.65  & 6.29  \\
    ProDA & 4.61  & 6.60  & 3.02  & 6.67  & 1.51  & 5.36  & 1.49  & 8.22  & 4.84  & 3.11  & 1.59  & 4.27  \\
    KgCoOp & 1.96  & 3.16  & 2.81  & 5.99  & 1.98  & 12.53  & 1.28  & 7.03  & 6.82  & 2.64  & 1.8   & 4.36  \\
    MaPLe & 1.33  & 2.51  & 2.02  & 13.43  & 1.26  & 10.38  & 3.01  & 15.71  & 9.06  & 2.38  & 2.39  & 5.77  \\
    ProGrad & 1.91  & 2.98  & 2.88  & 5.45  & 2.04  & 13.18  & 1.75  & 4.22  & 7.40  & 2.85  & 1.81  & 4.22  \\
    PromptSRC & 1.62  & 3.08  & 2.16  & 5.77  & 1.60  & 9.79  & 0.77  & 5.79  & 7.35  & 2.64  & 1.71  & 3.84  \\
    \bottomrule
    \end{tabular}%
    }
    \subcaption{w/o calibration}
    \end{subtable}

    \begin{subtable}{\linewidth}
        \centering
    \resizebox{1.0\textwidth}{!}{
    \begin{tabular}{ccccccccccccc}
    \toprule
    Methods & Caltech101  & OxfordPets & StanfordCars & Flowers102 & Food101 & FGVCAircraft & SUN397 & DTD   & EuroSAT & UCF101 & ImageNet & AVG \\
    \midrule
    ZeroshotCLIP & 1.60  & 3.42  & 3.31  & 4.91  & 1.83  & 6.55  & 3.48  & 6.86  & 9.12  & 5.52  & 2.09  & 4.43  \\
    CoOp  & 3.17  & 1.82  & 5.16  & 10.19  & 1.78  & 17.38  & 4.05  & 10.48  & 8.62  & 8.67  & 5.67  & 7.00  \\
    CoCoOp & 3.65  & 2.43  & 2.21  & 7.74  & 1.64  & 9.03  & 1.09  & 7.47  & 13.49  & 2.70  & 1.61  & 4.82  \\
    ProDA & 4.87  & 4.72  & 3.28  & 6.32  & 0.70  & 7.40  & 1.06  & 5.68  & 3.33  & 4.14  & 2.42  & 3.99  \\
    KgCoOp & 1.84  & 3.11  & 3.12  & 5.90  & 1.94  & 11.78  & 1.67  & 7.09  & 6.59  & 2.69  & 1.83  & 4.32  \\
    MaPLe & 1.26  & 2.51  & 2.75  & 11.28  & 1.50  & 9.06  & 1.22  & 8.16  & 8.55  & 2.30  & 2.11  & 4.61  \\
    ProGrad & 1.97  & 3.31  & 2.29  & 5.04  & 1.85  & 10.46  & 1.32  & 3.49  & 6.90  & 2.42  & 2.11  & 3.74  \\
    PromptSRC & 1.58  & 2.98  & 2.39  & 5.03  & 1.55  & 8.55  & 0.79  & 5.50  & 7.24  & 2.46  & 1.84  & 3.63  \\ 
    \bottomrule
    \end{tabular}%
    }
    \subcaption{Distance-Aware Calibration}
    \end{subtable}
    \label{tab_ece} 
\end{table}

\begin{table}[H]
    \centering
    \caption{Adaptive Calibration Error (ACE) on new classes in downstream datasets using various tuning methods for CLIP-ViT-B/16.}
    \begin{subtable}{\linewidth}
        \centering
    \resizebox{1.0\textwidth}{!}{
    \begin{tabular}{ccccccccccccc}
    \toprule
    Methods & Caltech101  & OxfordPets & StanfordCars & Flowers102 & Food101 & FGVCAircraft & SUN397 & DTD   & EuroSAT & UCF101 & ImageNet & AVG \\
    \midrule
    ZeroshotCLIP & 1.45  & 3.41  & 3.34  & 4.77  & 1.78  & 6.06  & 3.64  & 8.93  & 9.39  & 5.50  & 2.15  & 4.58  \\
    CoOp  & 3.46  & 1.66  & 12.47  & 18.34  & 3.83  & 28.44  & 13.93  & 26.86  & 12.71  & 19.18  & 10.69 & 13.78  \\
    CoCoOp & 3.30  & 2.02  & 2.30  & 11.49  & 1.65  & 11.19  & 2.00  & 15.73  & 12.68  & 4.38  & 1.48  & 6.20  \\
    ProDA & 4.36  & 6.59  & 3.13  & 6.87  & 1.39  & 5.58  & 1.68  & 8.10  & 5.16  & 3.37  & 1.57  & 4.35  \\
    KgCoOp & 1.56  & 3.05  & 2.78  & 6.79  & 1.99  & 12.48  & 1.52  & 7.34  & 6.83  & 2.61  & 1.73  & 4.43  \\
    MaPLe & 1.11  & 2.31  & 2.04  & 13.43  & 1.24  & 10.31  & 2.95  & 15.38  & 9.10  & 2.53  & 2.39  & 5.71  \\
    ProGrad & 1.22  & 2.86  & 3.13  & 5.90  & 2.04  & 12.96  & 1.65  & 5.18  & 7.66  & 2.55  & 1.84  & 4.27  \\
    PromptSRC & 1.59  & 2.95  & 2.34  & 6.05  & 1.54  & 9.67  & 0.99  & 5.75  & 7.54  & 2.98  & 1.75  & 3.92  \\
    \bottomrule
    \end{tabular}%
    }
    \subcaption{w/o calibration}
    \end{subtable}

    \begin{subtable}{\linewidth}
        \centering
    \resizebox{1.0\textwidth}{!}{
    \begin{tabular}{ccccccccccccc}
    \toprule
    Methods & Caltech101  & OxfordPets & StanfordCars & Flowers102 & Food101 & FGVCAircraft & SUN397 & DTD   & EuroSAT & UCF101 & ImageNet & AVG \\
    \midrule
    ZeroshotCLIP & 1.45  & 3.41  & 3.34  & 4.77  & 1.78  & 6.06  & 3.64  & 8.93  & 9.39  & 5.50  & 2.15  & 4.58  \\
    CoOp  & 2.60  & 1.70  & 5.17  & 10.18  & 1.75  & 17.27  & 4.00  & 10.51  & 8.58  & 8.63  & 5.66  & 6.91  \\
    CoCoOp & 2.97  & 2.27  & 2.35  & 7.69  & 1.62  & 8.92  & 1.09  & 7.73  & 13.49  & 2.66  & 1.69  & 4.77  \\
    ProDA & 4.64  & 4.72  & 3.37  & 6.54  & 0.64  & 7.73  & 1.04  & 5.64  & 3.99  & 4.19  & 2.37  & 4.08  \\
    KgCoOp & 1.56  & 3.02  & 3.11  & 6.61  & 1.93  & 11.74  & 1.82  & 7.26  & 6.63  & 2.70  & 1.78  & 4.38  \\
    MaPLe & 1.19  & 2.44  & 2.57  & 11.28  & 1.48  & 8.90  & 1.37  & 8.24  & 9.12  & 2.32  & 2.12  & 4.64  \\
    ProGrad & 1.21  & 3.19  & 2.28  & 5.66  & 1.88  & 10.29  & 1.42  & 3.82  & 7.10  & 2.20  & 2.12  & 3.74  \\
    PromptSRC & 1.65  & 2.86  & 2.34  & 5.50  & 1.48  & 8.46  & 1.03  & 5.42  & 7.43  & 2.54  & 1.83  & 3.69  \\
    \bottomrule
    \end{tabular}%
    }
    \subcaption{Distance-Aware Calibration}
    \end{subtable}
    \label{tab_ace} 
\end{table}

\begin{table}[h]
    \centering
    \caption{Maximum Calibration Error (MCE) on new classes in downstream datasets using various tuning methods for CLIP-ViT-B/16.}
    \begin{subtable}{\linewidth}
        \centering
    \resizebox{1.0\textwidth}{!}{
    \begin{tabular}{ccccccccccccc}
    \toprule
    Methods & Caltech101  & OxfordPets & StanfordCars & Flowers102 & Food101 & FGVCAircraft & SUN397 & DTD   & EuroSAT & UCF101 & ImageNet & AVG \\
    \midrule
    ZeroshotCLIP & 0.60  & 1.19  & 0.71  & 1.07  & 0.62  & 1.84  & 0.63  & 2.57  & 2.44  & 1.16  & 0.51  & 1.21  \\
    CoOp  & 2.60  & 0.62  & 2.40  & 5.46  & 1.43  & 5.00  & 4.36  & 7.08  & 4.42  & 6.09  & 2.67  & 3.83  \\
    CoCoOp & 2.18  & 0.97  & 0.59  & 3.08  & 0.55  & 2.61  & 0.66  & 3.79  & 3.81  & 1.13  & 0.42  & 1.80  \\
    ProDA & 3.56  & 1.66  & 0.65  & 1.34  & 0.61  & 1.44  & 0.40  & 1.72  & 1.42  & 0.89  & 0.33  & 1.27  \\
    KgCoOp & 0.63  & 1.25  & 0.78  & 1.46  & 0.65  & 3.41  & 0.34  & 1.51  & 1.96  & 0.55  & 0.42  & 1.18  \\
    MaPLe & 0.47  & 1.03  & 0.48  & 5.66  & 0.45  & 2.47  & 0.88  & 4.24  & 2.70  & 0.91  & 0.74  & 1.82  \\
    ProGrad & 0.60  & 1.10  & 1.05  & 1.28  & 0.61  & 3.16  & 0.60  & 1.32  & 2.40  & 0.72  & 0.55  & 1.22  \\
    PromptSRC & 0.66  & 1.02  & 0.51  & 1.2   & 0.56  & 2.41  & 0.23  & 1.59  & 2.65  & 0.65  & 0.54  & 1.09  \\
    \bottomrule
    \end{tabular}%
    }
    \subcaption{w/o calibration}
    \end{subtable}

    \begin{subtable}{\linewidth}
        \centering
    \resizebox{1.0\textwidth}{!}{
    \begin{tabular}{ccccccccccccc}
    \toprule
    Methods & Caltech101  & OxfordPets & StanfordCars & Flowers102 & Food101 & FGVCAircraft & SUN397 & DTD   & EuroSAT & UCF101 & ImageNet & AVG \\
    \midrule
    ZeroshotCLIP & 0.60  & 1.19  & 0.71  & 1.07  & 0.62  & 1.84  & 0.63  & 2.57  & 2.44  & 1.16  & 0.51  & 1.21  \\
    CoOp  & 1.50  & 0.66  & 1.21  & 2.08  & 0.49  & 3.57  & 1.03  & 2.24  & 3.03  & 1.70  & 1.28  & 1.71  \\
    CoCoOp & 1.92  & 1.06  & 0.55  & 1.60  & 0.53  & 2.34  & 0.28  & 1.75  & 4.41  & 0.58  & 0.43  & 1.40  \\
    ProDA & 3.98  & 1.27  & 0.81  & 1.20  & 0.28  & 1.65  & 0.42  & 1.61  & 1.13  & 1.32  & 0.81  & 1.32  \\
    KgCoOp & 0.62  & 1.21  & 0.85  & 1.37  & 0.64  & 3.02  & 0.42  & 1.33  & 1.75  & 0.77  & 0.41  & 1.13  \\
    MaPLe & 0.39  & 1.06  & 0.77  & 4.42  & 0.50  & 2.26  & 0.40  & 1.95  & 2.55  & 0.66  & 0.64  & 1.42  \\
    ProGrad & 0.71  & 1.17  & 0.88  & 1.18  & 0.58  & 2.60  & 0.40  & 1.18  & 1.96  & 0.69  & 0.6   & 1.09  \\
    PromptSRC & 0.64  & 0.96  & 0.58  & 1.23  & 0.55  & 2.31  & 0.22  & 1.33  & 2.72  & 0.79  & 0.57  & 1.08  \\
    \bottomrule
    \end{tabular}%
    }
    \subcaption{Distance-Aware Calibration}
    \end{subtable}
    \label{tab_mce} 
\end{table}

\begin{table}[H]
    \centering
    \caption{Proximity-Informed Expected Calibration Error (PIECE) on new classes in downstream datasets using various tuning methods for CLIP-ViT-B/16.}
    \begin{subtable}{\linewidth}
        \centering
    \resizebox{1.0\textwidth}{!}{
    \begin{tabular}{ccccccccccccc}
    \toprule
    Methods & Caltech101  & OxfordPets & StanfordCars & Flowers102 & Food101 & FGVCAircraft & SUN397 & DTD   & EuroSAT & UCF101 & ImageNet & AVG \\
    \midrule
    ZeroshotCLIP & 3.95  & 3.88  & 4.79  & 7.62  & 2.17  & 10.49  & 4.31  & 13.75  & 10.37  & 7.72  & 3.05  & 6.55  \\
    CoOp  & 6.21  & 4.26  & 12.73  & 18.80  & 4.07  & 29.03  & 14.01  & 27.29  & 15.07  & 19.79  & 10.71 & 14.72  \\
    CoCoOp & 5.49  & 2.96  & 4.54  & 12.35  & 2.19  & 13.52  & 3.46  & 18.64  & 15.08  & 7.68  & 2.54  & 8.04  \\
    ProDA & 6.08  & 6.89  & 5.56  & 8.90  & 2.39  & 8.87  & 2.88  & 12.84  & 7.96  & 7.46  & 2.48  & 6.57  \\
    KgCoOp & 4.14  & 3.86  & 4.95  & 8.81  & 2.33  & 14.87  & 3.08  & 12.17  & 9.27  & 7.16  & 2.73  & 6.67  \\
    MaPLe & 3.39  & 3.27  & 4.53  & 14.18  & 1.90  & 12.33  & 3.93  & 18.35  & 11.95  & 6.51  & 3.17  & 7.59  \\
    ProGrad & 4.45  & 3.43  & 5.29  & 8.31  & 2.49  & 15.21  & 3.33  & 10.64  & 10.95  & 7.15  & 2.96  & 6.75  \\
    PromptSRC & 3.80  & 3.82  & 4.41  & 8.02  & 2.07  & 12.34  & 2.81  & 12.10  & 10.31  & 6.49  & 2.64  & 6.26  \\
    \bottomrule
    \end{tabular}%
    }
    \subcaption{w/o calibration}
    \end{subtable}

    \begin{subtable}{\linewidth}
        \centering
    \resizebox{1.0\textwidth}{!}{
    \begin{tabular}{ccccccccccccc}
    \toprule
    Methods & Caltech101  & OxfordPets & StanfordCars & Flowers102 & Food101 & FGVCAircraft & SUN397 & DTD   & EuroSAT & UCF101 & ImageNet & AVG \\
    \midrule
    ZeroshotCLIP & 3.95  & 3.88  & 4.79  & 7.62  & 2.17  & 10.49  & 4.31  & 13.75  & 10.37  & 7.72  & 3.05  & 6.55  \\
    CoOp  & 5.89  & 4.15  & 6.84  & 12.07  & 2.58  & 19.43  & 4.92  & 13.13  & 11.88  & 12.23  & 6.08  & 9.02  \\
    CoCoOp & 5.37  & 3.19  & 4.47  & 10.14  & 2.14  & 12.19  & 2.88  & 12.78  & 15.69  & 7.11  & 2.65  & 7.15  \\
    ProDA & 6.25  & 5.24  & 5.61  & 8.80  & 1.59  & 10.06  & 2.74  & 11.44  & 7.26  & 7.61  & 3.20  & 6.35  \\
    KgCoOp & 4.13  & 3.81  & 5.05  & 8.73  & 2.28  & 14.47  & 3.34  & 12.21  & 9.13  & 7.03  & 2.71  & 6.63  \\
    MaPLe & 3.59  & 3.33  & 4.90  & 12.40  & 2.12  & 11.77  & 2.88  & 14.09  & 11.96  & 6.70  & 2.99  & 6.98  \\
    ProGrad & 4.77  & 3.69  & 5.00  & 8.07  & 2.42  & 13.64  & 3.12  & 10.86  & 10.40  & 6.90  & 3.15  & 6.55  \\
    PromptSRC & 3.75  & 3.74  & 4.48  & 8.00  & 1.98  & 11.71  & 2.84  & 12.41  & 10.10  & 6.08  & 2.74  & 6.17  \\
    \bottomrule
    \end{tabular}%
    }
    \subcaption{Distance-Aware Calibration}
    \end{subtable}
    \label{tab_piece} 
\end{table}

\newpage

\end{document}